\documentclass[%
 aip,
 amsmath,amssymb,
reprint,%
]{revtex4-2}

\usepackage{graphicx}
\usepackage{dcolumn}
\usepackage{bm}
\usepackage[colorlinks=true,allcolors=black]{hyperref}

\usepackage[utf8]{inputenc}
\usepackage[T1]{fontenc}
\usepackage{mathptmx}
\DeclareMathAlphabet{\mathcal}{OMS}{cmsy}{m}{n}
\usepackage{etoolbox}
\usepackage{color} 

\usepackage{amsmath}
\usepackage{amsthm}
\usepackage{physics}
\usepackage{mathtools}
\DeclarePairedDelimiter\floor{\lfloor}{\rfloor}
\usepackage{url}
\usepackage{comment}
\usepackage{tabularx}
\usepackage{multirow}

\usepackage[capitalise]{cleveref}

\newtheorem{theorem}{Theorem}[section]
\newtheorem{result}[theorem]{Result}

\newtheorem{remark}[theorem]{Remark}
\newtheorem{example}[theorem]{Example}

\newcommand{\x}{\mathbf{x}}
\newcommand{\X}{\mathbf{X}}
\newcommand{\y}{\mathbf{y}}

\newcommand{\rs}{\mathbf{r}}
\newcommand{\R}{\mathbf{R}}
\newcommand{\us}{\mathbf{u}}
\newcommand{\U}{\mathbf{U}}
\newcommand{\vs}{\mathbf{v}}
\newcommand{\V}{\mathbf{V}}
\newcommand{\F}{\mathbf{F}}
\newcommand{\f}{\mathbf{f}}

\newcommand{\g}{\mathbf{g}}
\newcommand{\W}{\mathbf{W}}
\newcommand{\w}{\mathbf{w}}
\newcommand{\co}{\mathbf{c}}
\newcommand{\C}{\mathbf{C}}
\newcommand{\Ntrain}{N_{\mathrm{train}}}
\newcommand{\Ntest}{N_{\mathrm{test}}}

\newcommand{\re}[1]{{\color{black} #1}}
\newcommand{\topic}[1]{{\noindent \textbf{#1}.}}

\usepackage{lipsum}

\DeclareMathOperator*{\argmin}{arg\,min} 

\makeatletter
\def\@email#1#2{%
 \endgroup
 \patchcmd{\titleblock@produce}
  {\frontmatter@RRAPformat}
  {\frontmatter@RRAPformat{\produce@RRAP{*#1\href{mailto:#2}{#2}}}\frontmatter@RRAPformat}
  {}{}
}%
\makeatother
\begin{document}

\preprint{AIP/123-QED}

\title[On the emergence of numerical instabilities in Next Generation Reservoir Computing]{On the emergence of numerical instabilities in Next Generation Reservoir Computing}

\author{Edmilson Roque dos Santos}
\email{edmilson.roque.usp@gmail.com}
\affiliation{ 
Department of Electrical and Computer Engineering, Clarkson
University, Potsdam, 13699, New York, U.S.A.
}%
\affiliation{Clarkson Center for Complex Systems Science, Clarkson
University, Potsdam, 13699, New York, U.S.A.
}%

\author{Erik Bollt}
\affiliation{ 
Department of Electrical and Computer Engineering, Clarkson
University, Potsdam, 13699, New York, U.S.A.
}%
\affiliation{Clarkson Center for Complex Systems Science, Clarkson
University, Potsdam, 13699, New York, U.S.A.
}%

\date{\today}

\begin{abstract}
Next Generation Reservoir Computing (NGRC) is a low-cost machine learning method for forecasting chaotic time series from data. {Computational efficiency is crucial for scalable reservoir computing, requiring better strategies to reduce training cost.} In this work, we uncover a connection between the numerical conditioning of the NGRC feature matrix --- formed by polynomial evaluations on time-delay coordinates --- and the long-term NGRC dynamics. {We show that NGRC can be trained without regularization, reducing computational time. Our contributions are twofold. First,} merging tools from numerical linear algebra and ergodic theory of dynamical systems, we systematically study how the feature matrix conditioning varies across hyperparameters. We demonstrate that the NGRC feature matrix tends to be ill-conditioned for short time lags, high-degree polynomials, and short length of training data. {Second}, we evaluate the impact of different numerical algorithms (Cholesky, singular value decomposition (SVD), and lower-upper (LU) decomposition) for solving the regularized least-squares problem. {Our results reveal that SVD-based training achieves accurate forecasts without regularization, being preferable when compared against the other algorithms.}
\end{abstract}

\maketitle

\begin{quotation}
Next Generation Reservoir Computing (NGRC) is a reservoir computing (RC) variant characterized by a reduced number of hyperparameters, data efficiency and forecasting quality, and easy interpretability. Despite its success in forecasting tasks, a central challenge remains: ensuring the dynamical stability of the model over long-term forecasts {in a computationally efficient manner}. Recent theoretical work points out a subtle interplay between approximation error and overfitting, yet no analytical method currently exists for selecting hyperparameters that guarantee the NGRC dynamical stability. We investigate the connection between the choice of hyperparameters and the condition number of the NGRC feature matrix, formed by polynomial evaluations on time-delay coordinates. Notably, the NGRC model features structured matrices, such as Vandermonde-like and Hankel-like matrices, which have been extensively studied in numerical analysis. Merging numerical linear algebra and ergodic theory of dynamical systems, we characterize the conditioning of the feature matrix in terms of the maximum degree of the polynomial basis, the interplay between delay dimension and time lag, and the length of training data. {This allows us to guide better hyperparameter choices so that an accurate NGRC model can be trained without any regularization, saving computational time.} This numerical analysis is a first step toward designing better strategies to select hyperparameters in reservoir computing.
\end{quotation}


\section{Introduction}


Forecasting time series is a fundamental problem across scientific disciplines \cite{CASDAGLI1989335,MARCELLINO2006499,McElroy_2015}, especially when dealing with chaotic systems time series \cite{Kantz_Schreiber_2003}. In cases where the governing equations are unknown, the task becomes significantly more challenging. Reservoir Computing (RC) has emerged as a powerful machine learning method for forecasting chaotic dynamical systems directly from data \cite{Jaeger_2004,Lu_2017,Lu_2018,Pathak_2017,Pathak_2018}. A recent promising variant, the Next Generation Reservoir Computing (NGRC) model, recasts the input data as a nonlinear vector autoregressive model \cite{Gauthier2021}. It requires fewer hyperparameters, shorter training times, and less ``warm-up'' compared to traditional RC methods \cite{Barbosa_2022,Zhang_2023}. NGRC stands out for its lower computational cost for forecasting but also other tasks, such as control \cite{Kent_2024}, reconstruction from partial measurements \cite{Gauthier2021,Ratas_2024}, basin reconstruction \cite{Gauthier_2022_basins,Zhang_2023}, and experimental implementation \cite{Kent_2024_hardware,Cox_2024,Wang_2024}.  


Despite its success in forecasting chaotic time series, a central problem is {to ensure computational efficiency for} the reservoir dynamical stability for short- and long-term prediction \cite{Lukosevicius_2012,Lu_2018,Zhang_2023,grigoryeva2025_forecasting_causal_dynamics}. Traditionally, in NGRC, training consists of {fixing a choice of hyperparameter and} solving a regularized least squares minimization problem (Tikhonov regularization or {ridge regression}) to find the readout weight matrix that best predicts one step into the future. The trained model is then run autonomously for forecasting. 
{The successful NGRC model should have a good forecast, not diverging to infinity in the long run. This process must be repeated for each hyperparameter, potentially adding computational cost.}

Recent theoretical work links reservoir dynamical stability to properties of the original dynamics and the approximation capacity used during training \cite{Sauer_1998_spurious_Lyapunov_spectrum,DECHERT_200059,Berry_2023,Berry_2023,grigoryeva2025_forecasting_causal_dynamics}. By seeing reservoirs as high-dimensional embeddings \cite{HART_2020,Grigoryeva_2023}, the largest Lyapunov exponent of the reservoir dynamics should not differ much from the original dynamics. However, there is no analytical method to guide hyperparameter selection that guarantees NGRC stability. Developing an analytical method for selecting hyperparameters requires understanding how these choices influence the NGRC dynamics. The mapping from hyperparameters to NGRC dynamics is often nontrivial. For instance, adjusting only the regularization strength \cite{Lu_2018} can drastically alter the reservoir's behavior. Moreover, an unstable NGRC model can emerge when the regularizer parameter is not scaled to the length of the training data \cite{Zhang_2025}. 

A promising direction comes from numerical linear algebra: for each hyperparameter configuration, we can examine the condition number of the feature matrix --- the matrix that evaluates the input trajectory along the polynomial functions of time-delayed coordinates during training. The condition number quantifies the sensitivity of the least squares solution to perturbations \cite{GoluVanl96,trefethen1997numerical}. When the matrix is ill-conditioned (large condition number), even small changes in the data can produce large readout weights. These large weights are then fed back into the reservoir, potentially driving the model into dynamical instability \cite{Zhang_2025}. {Moreover, ill-conditioning potentially requires regularization, adding computational time to find the regularizer parameter that best forecasts during testing.} 

In this paper, we exploit a key structural insight of NGRC: its architecture naturally connects concepts from numerical linear algebra and dynamical systems theory. We show that the NGRC feature matrix possesses well-known structured forms — specifically, a Vandermonde-like matrix, due to the evaluation of polynomial basis functions on time series data, similarly done in polynomial interpolation; and a Hankel-like matrix, due to the use of time-delay coordinates. These matrix classes have been rigorously analyzed in numerical analysis, particularly in terms of their conditioning properties \cite{GoluVanl96,trefethen1997numerical,Higham_2002,Beckermann_Townsend_2017}. By leveraging this structure, we systematically characterize how the conditioning of the feature matrix varies with hyperparameter choices, and we pinpoint the regimes that lead to severe ill-conditioning. Ill-conditioned matrices are a source of numerical instabilities, depending on the condition number and the residual error \cite{GoluVanl96,trefethen1997numerical,Higham_2002}. We {study chaotic systems} to demonstrate that the NGRC feature matrix becomes ill-conditioned, particularly for short time lags, high-degree polynomials, and short training lengths. {Using this information, we guide better choices of hyperparameters that allow for NGRC training without any regularization. Furthermore,} we test different numerical algorithms to solve the regularized least squares problem of the NGRC training: Cholesky, singular value decomposition (SVD), and lower-upper (LU) decomposition. {Our findings show that SVD-based training is more stable and preferable than the other two algorithms.} 

This paper is organized as follows: the preliminaries are presented in \cref{sec:notation}. We then introduce the Lorenz system generated by the explicit forward Euler model as a toy model in \cref{sec:key_example_explicit} to motivate the numerical analysis perspective in \cref{sec:perturbation_bounds,sec:num_algs,sec:numerical_algo_behavior}, and problem formulation in \cref{sec:problem_statement}. Then, \cref{sec:condition_number_characterization} characterizes the condition number of the feature matrix. Subsequently, \cref{sec:condition_guides_hyperparameter} demonstrates two relevant application scenarios where a better choice of hyperparameters allows for an accurate NGRC model without regularization. Finally, we summarize the results and present the discussion in \cref{sec:discussion_conclusions}. 

\section{Preliminaries} 

\subsection{Notation}
\label{sec:notation}
A {$d-$dimensional vector} is denoted as $\us = (u_1, \dots, u_d)$, whereas for a collection of $d-$dimensional vectors, we utilize double indices: $\us_i = (u_{i,1}, u_{i, 2}, \dots, u_{i,d})$. The operation $\mathbf{vec}$ concatenates vectors into a single vector, and $\top$ is the transpose. Let $\|\cdot\|$ and $\|\cdot\|_2$ be the Euclidean and spectral norm, respectively. The spaces $L^{1}(\mu)$ and $L^{2}(\mu)$ correspond to the space of integrable and square-integrable functions with respect to a probability measure $\mu$, respectively. We denote $\mathbf{1}_d$ as the $d \times d$ identity matrix. We utilize Landau's notation $\mathcal{O}(\varepsilon)$ denoting a function for which there exists a positive constant $K$ such that $0 \leq |\mathcal{O}(\varepsilon)| \leq K \varepsilon$ for the limit $\varepsilon \to 0$ or $\varepsilon \to \infty$, being clear by the context. 

\subsection{Learning dynamics}
\label{sec:discrete_map}
We assume a dynamical system exists $\f: M \to M$ lying on a compact $d-$dimensional metric space $M \subset \mathbb{R}^d$. For our purposes, $\f$ is given by discretizing an ordinary differential equation sampled uniformly at every $h$ time step. Instead of knowing $\f$, we only have access to a trajectory $\{\x_n\}_{n \geq 0}$ for a given initial condition $\x_0$, where $\x_n = \f^{n}(\x_0)$, where $\f^n$ denotes the $n$-fold composition of the map $\f$. {The observed time series is normalized coordinate-wise by applying the transformation $\x_{n, i} \mapsto \frac{\x_{n, i}}{\max_n\{\x_{n, i}\}}$ for each coordinate $i = 1, \dots, d$. The problem is to forecast the original dynamics forward in time only from the observed data.}

\subsection{Next Generation Reservoir Computing}
\label{sec:NGRC} 
We focus on using Next Generation Reservoir Computing (NGRC) \cite{Gauthier2021}, which recasts the input data as a nonlinear vector autoregression model, akin to a numerical integration scheme. {This particular architecture originated from the mathematical equivalence between a traditional RC and the NGRC: there exists an NVAR that performs equally as well as an optimized RC \cite{Bollt_2021}. The key observation is that RC can be recast using a linear activation function and nonlinear readouts, akin to a Volterra series, which have been proven to be a universal approximator \cite{Boyd_2003_fading_control}. Although this equivalence only occurs for infinite memory, in practice, truncated finite memory suffices, speeding up training and making it the main advantage of the NGRC. \re{Interestingly, NGRC has also been shown to be a special case of polynomial kernel regression using time-delayed coordinates \cite{Grigoryeva_2025_infinite_dimensional_NGRC}, showing its relevance inside the broader context of machine learning approaches to dynamical systems.} }  

As the original formulation, the NGRC model utilizes a linear combination of polynomial functions evaluated at the current and time-delayed coordinates. More precisely, fix a delay dimension $k \in \mathbb{N}$ and time lag (or time skip) $\tau \in \mathbb{N}$, and define the embedding map $\g_{k, \tau}:M \to (\mathbb{R}^d)^k$ by
\begin{align}\label{eq:embedding_map}
    \g_{k, \tau}(\x) = \mathbf{vec}\Big(\x, \f^{\tau}(\x), \dots, \f^{(k - 1) \tau}(\x)\Big),
\end{align}
which is valid for any point $\x \in M$, and the coordinate in $(\mathbb{R}^d)^k$ is denoted as $\X = \g_{k, \tau}(\x)$. Specifically, for the time series $\{\x_n\}_{n \geq 0}$, the following holds:
\begin{align*}
    \X_n \equiv \g_{k,\tau}(\x_n) &:= \mathbf{vec}\Big(\x_n, \f^{\tau}(\x_n), \dots, \f^{(k - 1) \tau}(\x_n)\Big) \\
    &= \mathbf{vec}\Big(\x_n, \x_{n + \tau}, \dots, \x_{n + (k - 1) \tau}\Big).
\end{align*}
Let $\alpha \in (\mathbb{N}^{d})^k$ be the multi-index {vector} with $|\alpha| = \sum_{j = 0}^{k - 1} \sum_{i = 1}^d \alpha_{i, j}$. The set of multivariate monomials in $kd$ variables up to degree $p$ is given by:
\begin{align*}
    \mathcal{P}_p^{kd} = \Big\{
     \X_n^{\alpha} = \prod_{j = 0}^{k - 1} \prod_{i = 1}^{d} x_{n + j\tau, i}^{\alpha_{i, j}}: \quad |\alpha| \leq p \Big\},
\end{align*}
and has cardinality $m := \binom{kd + p}{p}$. For example, if $k=1$, $d=2$, and $p=2$, then $\mathcal{P}_2^{2}$ includes terms like $x_{n, 1}$, $x_{n, 1}^2$, $x_{n, 2}$, $x_{n, 2}^2$, $x_{n,1}x_{n,2}$, and constant $1$. This induces a high-dimensional map $\psi: (\mathbb{R}^d)^k \to \mathbb{R}^{m}$ as
\begin{align}\label{eq:high_dim_phi}
    \psi(\X_n) = \mathbf{vec}\Big(\X_n^{\alpha}\Big)_{|\alpha| \leq p},
\end{align}
which constructs a $m$-dimensional vector by the collection of monomials $\X_n^{\alpha}$. 

\topic{Training phase} The NGRC model assumes that the time series can be recast akin as an iterative scheme induced by the high-dimensional map $\psi$ 
\begin{align}\label{eq:NGRC}
    \x_{n + 1} = \x_{n} + \W \psi(\X_{n - (k - 1)\tau}),\quad n \geq (k - 1)\tau. 
\end{align}
The goal of the training phase is to determine the readout matrix $\W \in \mathbb{R}^{d \times m}$ that best satisfies the above relation at each time step. The available time series $\{\x_n\}_{n \geq 0}^{\Ntrain + (k - 1)\tau}$ has the number of points necessary to use \cref{eq:NGRC}, where $\Ntrain$ corresponds to the number of training data points and the warm-up $(k-1)\tau$ is the required number of data points to construct the embedded coordinate $\X_n$.  

For the case of a sufficiently long time series ($\Ntrain > m$), finding $\W$ can be solved by a least squares with a Tikhonov regularization (or {ridge regression}). Let us write explicitly the rows of $\W$, such that $\W = (\w_1^{\top}, \dots, \w_d^{\top})$ where $\w_i$ is a $m$ dimensional vector. The regression can be made separately for each coordinate of the input trajectory as follows: introduce for $i$-th coordinate
\begin{align}\label{eq:diff_y}
    \y_i = \begin{pmatrix}
        x_{(k-1)\tau + 1, i} - x_{(k-1)\tau, i}\\
        x_{(k-1)\tau +2, i} - x_{(k-1)\tau + 1, i} \\
        \vdots \\
        x_{(k-1)\tau +\Ntrain, i} - x_{(k-1)\tau + \Ntrain - 1, i}
    \end{pmatrix}
\end{align}
and construct the $\Ntrain \times m$ feature matrix using $\psi(\X_n)$ as row vectors 
\begin{align}\label{eq:library_matrix}
    \Psi = \frac{1}{\sqrt{\Ntrain}} \begin{pmatrix}
        \psi^{\top}(\X_0)\\
        \psi^{\top}(\X_1) \\
        \vdots \\
        \psi^{\top}(\X_{\Ntrain - 1})
    \end{pmatrix}.
\end{align}
The solution to the following minimization problem yields each row of $\W$
\begin{align}\label{eq:w_out_training}
    \w_i(\beta) = \argmin_{\us \in \mathbb{R}^{m}} \Big\{ \|\y_i - \Psi \us\|^2 + \beta \|\us\|^2\Big\},
\end{align}
where $\beta$ is the regularizer parameter. This is equivalent to solving $d$ independent regressions. The unique solution to this minimization problem is 
\begin{align}\label{eq:solution_Tikhonov}
    \w_i(\beta) = (\Psi^{\top} \Psi + \beta \mathbf{1}_m)^{-1}\Psi^{\top} \y_i,
\end{align}
where $\mathbf{1}_m$ is the $m \times m$ identity matrix. 

\topic{Testing phase} The NGRC model then evolves autonomously according to the learned dynamics: 
\begin{align}\label{eq:autonomously_NGRC}
    \rs_{n + 1} &= \rs_{n} + \frac{1}{\sqrt{\Ntrain}} \W \psi(\R_n), \quad n \geq 0,
\end{align}
where $\rs \in \mathbb{R}^d$ denotes the $d-$dimensional state variable of the NGRC model,
\begin{align*}
    \R_n = \mathbf{vec}\Big(\rs_n, \rs_{n + \tau}, \dots, \rs_{n + (k - 1) \tau}\Big),
\end{align*}
and the autonomous evolution is initialized using the last $(k - 1)\tau$ values from the training set to construct the initial embedded coordinate given by 
\begin{align*}
    \rs_{n} &= \x_{\Ntrain + n}, \quad n = 0, 1, \dots, (k - 1) \tau - 1.
\end{align*}
The goal is that the NGRC trajectory $\{\rs_n\}_{n \geq 0}$ predicts features from the unseen trajectory of the original dynamics. We denote the number of iterations during the testing phase as $\Ntest$.

\begin{figure*}[t!]
\centering
\includegraphics[width=0.8\linewidth]{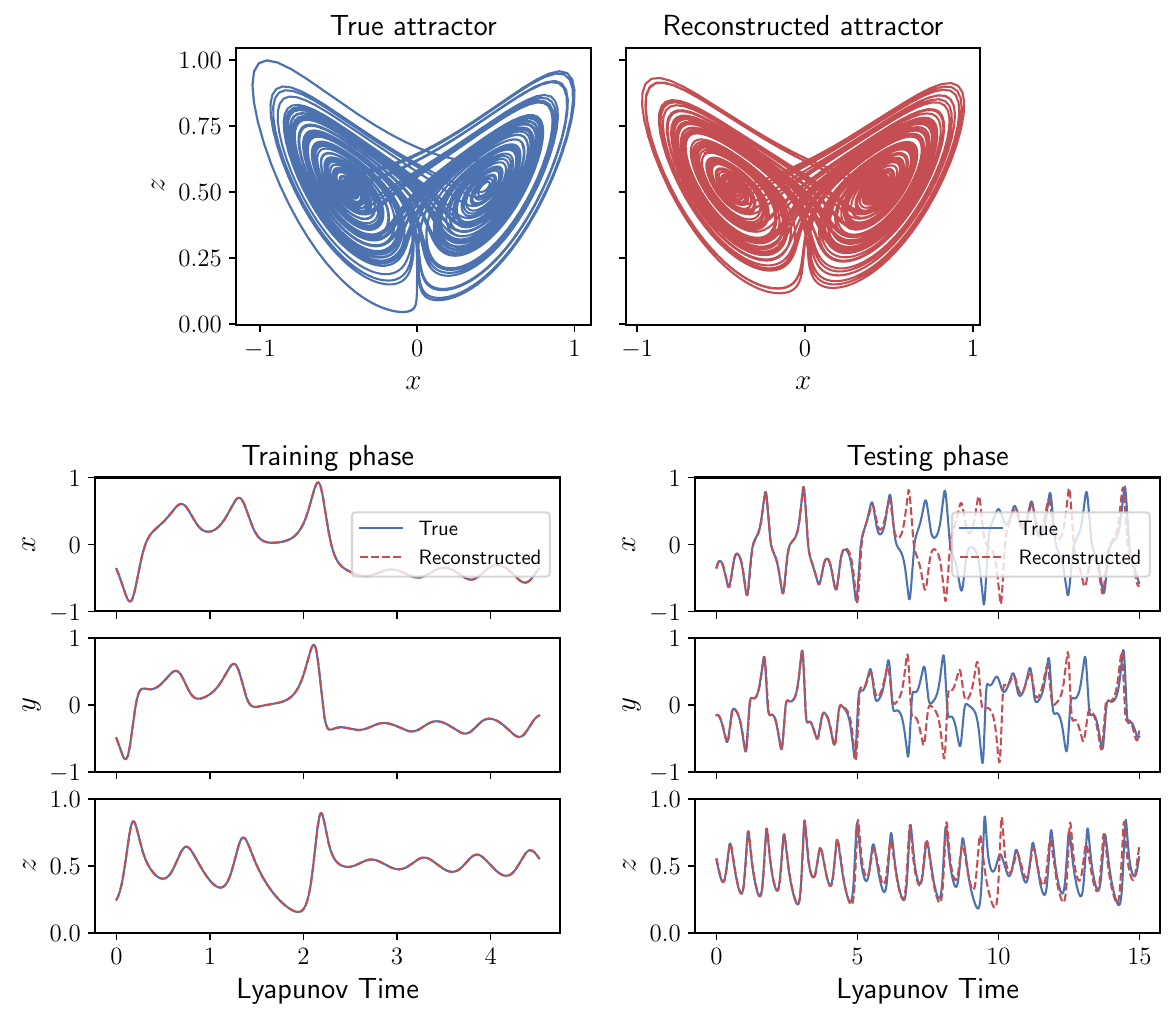}
\caption{\textbf{NGRC model accurately reproduces the Lorenz attractor.} The top panel shows the NGRC reconstruction (red) of the Lorenz attractor (blue). The bottom panel shows the NGRC model's performance over the training and testing phases. The horizontal axis shows time in Lyapunov time $\frac{1}{\Lambda}$ where $\Lambda = 0.9056$ is the maximum Lyapunov exponent of the Lorenz system. The parameters are $h = 0.01$, delay dimension $k = 1$, maximum degree $p = 2$, {time lag $\tau = 1$}, $\Ntrain = 500$, $\Ntest = 10000$ (but a smaller time window is shown), and regularizer parameter $\beta = 0$.}
\label{fig:NGRC_k_1_dt_001}
\end{figure*}

\subsection{Key example: Lorenz system with explicit forward Euler method} 
\label{sec:key_example_explicit}

A widely studied chaotic system often used as a benchmark in forecasting tasks for machine learning algorithms is the Lorenz 63 system \cite{Lorenz_1963}:
\begin{align}\label{eq:Lorenz_equation}
\begin{split}
    \dot{x} &= 10 (y - x) \\
    \dot{y} &= x(28 -  z) - y \\
    \dot{z} &= x y - \frac{8}{3} z.
\end{split}
\end{align}
Consider the time series given by integration of the Lorenz system vector field $\F$ using the explicit forward Euler method:
\begin{align}\label{eq:Euler_method}
    \x_{n + 1} = \x_{n} + h\F(\x_n),
\end{align}  
which defines a discrete-time dynamical system that approximates the continuous flow of the original dynamics for small $h$. Let us consider the NGRC model with $k = 1$ (absence of any time-delayed coordinates). Figure \ref{fig:NGRC_k_1_dt_001} shows the attractor reconstruction and forecasting capability of the NGRC model for the Lorenz system \cref{eq:Lorenz_equation}. The NGRC model captures the original dynamics, successfully predicting the trajectory for up to five Lyapunov times. This also extends to other dynamical features of the original dynamics. 

\begin{figure*}[t!]
\centering
\includegraphics[width=1.0\linewidth]{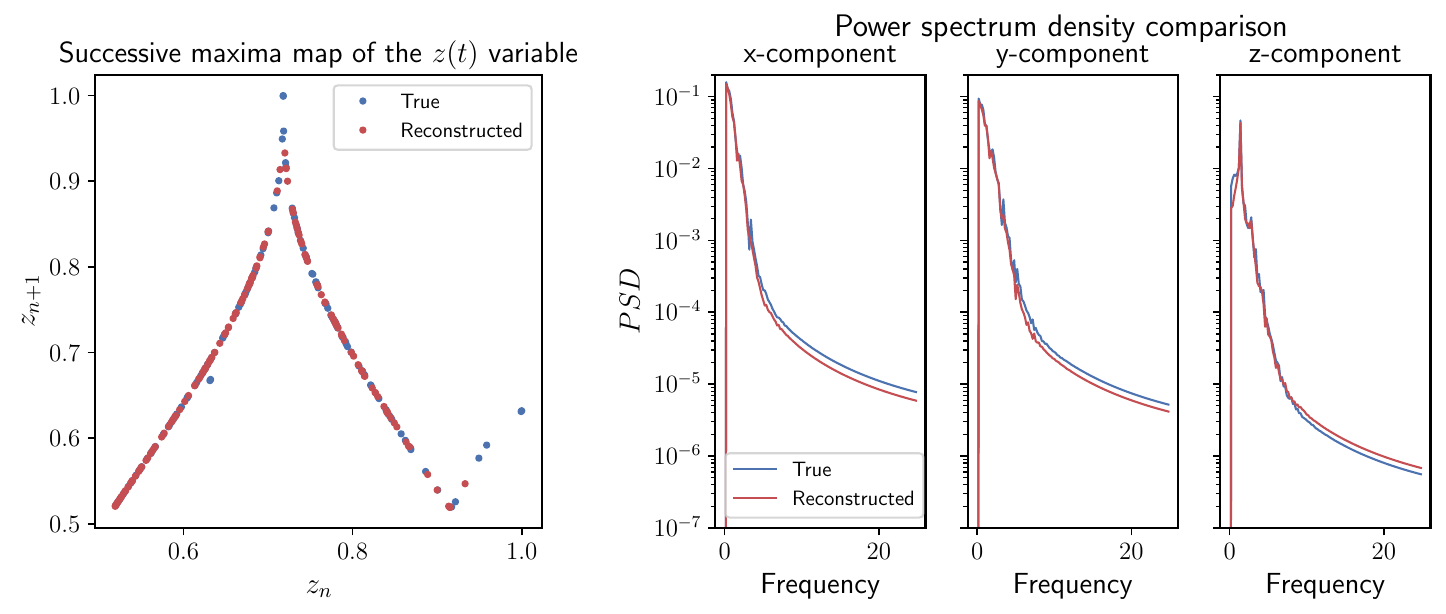}
\caption{\textbf{NGRC model captures topological and statistical features of the original dynamics.} The left panel displays the Poincaré return map of the successive local maxima of the original dynamics (in blue) and the reconstructed one (in red). The right panel depicts the power spectrum density of the original dynamics (in blue) and the reconstructed one (in red). Both curves mostly overlap for all frequencies, indicating that the NGRC model captures correctly the long-term statistics on the attractor. The parameters are the same used for \cref{fig:NGRC_k_1_dt_001}.}
\label{fig:NGRC_k_1_dt_001_tot_stat}
 \end{figure*}

{Edward Lorenz} realized that extracting successive local maxima of the $z-$coordinate yields a one-dimensional return map, which is referred to as the successive local maxima map, that is topologically conjugated to the tent map \cite{Lorenz_1963}. The left panel in \cref{fig:NGRC_k_1_dt_001_tot_stat} shows that the NGRC model successfully captures this map.  The right panel of \cref{fig:NGRC_k_1_dt_001_tot_stat} displays the power spectrum density of the three variables for the original dynamics and NGRC model, showing that the NGRC model also reproduces the long-term statistics of the Lorenz attractor. These observations demonstrate NGRC's ability to recover the chaotic dynamics using only observed trajectory data, without explicit knowledge of the underlying equations.

\subsection{Perturbation bounds for the least squares}
\label{sec:perturbation_bounds}
This example is insightful when we examine the numerical accuracy of the least squares problem involved during the NGRC training. In this setup, the NGRC model trained with $k = 1$ effectively learns an explicit forward Euler scheme, reducing to identifying the coefficients of a polynomial vector field from the time series data. For an arbitrary delay dimension $k$, using $\mathcal{P}_p^{kd}$ and the corresponding embedded vector $\X_n$, the explicit forward Euler discretization of a polynomial vector field implies that there exists a coefficient matrix $\C \in \mathbb{R}^{d \times m}$ such that:
\begin{align*}
    \x_{n + 1} = \x_{n} + \C \psi(\X_{n - (k - 1)\tau}),
\end{align*}
where $\C = (\co_1^{\top}, \dots, \co_d^{\top})$ and each row vector $\co_i^{\top}$ corresponds to the coefficient vector of the $i$-th coordinate of the vector field in the polynomial basis, where its magnitude depends on the time step $h$ \footnote{Since the explicit forward Euler method evaluates the vector field at current state only, the true coefficients $\co_i$ have zero entries corresponding to monomials functions evaluated at any of the time-delayed coordinates.}. Since the problem can be solved independently for each coordinate $i$, let us temporally drop the $i-$th dependence. If $\Psi$ is of full column rank, then the least squares solution ($\beta = 0$) is unique and is attained by $\co$:
\begin{align*}
    \co = \argmin_{\us \in \mathbb{R}^{m}} \big\{ \|\y - \Psi \us\|^2\big\},
\end{align*}
where $\y$ is the vector in \cref{eq:diff_y}. However, in practice, numerical linear algebra is required to assess the accuracy of the computed solution. As a result, the NGRC training corresponds to solving a perturbed least squares problem of the form
\begin{align}\label{eq:linear_equation}
     \w = \argmin_{\us \in \mathbb{R}^{m}} \big\{ \|\y + \Delta \y - (\Psi + \Delta \Psi) \us\|^2\big\},
\end{align}
where $\Delta \y$ and $\Delta \Psi$ are small perturbation errors (e.g., due to rounding errors and discretization cumulative error). The sensitivity of the least squares problem describes how small perturbations in $\y$ and $\Psi$ affect the solution. The numerical accuracy in recovering $\co$ heavily depends on two factors: 
\begin{itemize}
    \item The condition number of $\Psi$, $\kappa(\Psi) = \|\Psi\|_2\|\Psi^{\dagger}\|_2$, where $\Psi^{\dagger}$ is the pseudoinverse, or equivalently, $\kappa(\Psi):= \frac{\sigma_1}{\sigma_m}$ written in terms of the maximum and minimum singular values, $\sigma_1$ and $\sigma_m$, respectively.
    \item The magnitude of the minimum residual $\|\y - \Psi \w\|$, measuring how well the model fits the data. This introduces the closeness of fit $\theta = \sin^{-1}\big({\frac{\|\y - \Psi \w\|}{\|y\|}}\big)$. This angle characterizes the geometric misalignment between the $\y$ and the range of the feature matrix \cite{trefethen1997numerical}.
\end{itemize}
The following first-order bound quantifies the numerical accuracy quantified by the relative error \cite[Theorem 5.3.1]{GoluVanl96}: for $\varepsilon := \max \Big\{ \frac{\|\Delta \Psi\|}{\|\Psi\|},\frac{\|\Delta \y\|}{\|\y\|}\Big\}$ such that $\varepsilon \kappa(\Psi) < 1$ and $\sin(\theta) \neq 1$:
\begin{align}\label{eq:rel_error_kappa}
    \frac{\|\w - \co\|}{\|\co\|} \leq \varepsilon\Big( \frac{2 \kappa(\Psi)}{\cos(\theta)} + \tan{\theta} \kappa(\Psi)^2\Big) + \mathcal{O}(\varepsilon^2).
\end{align}
Even when $\Psi$ has full column rank, a large condition number (i.e., $\Psi$ is ill-conditioned) can cause the least squares solution to deviate drastically due to the perturbation. More precisely, $\varepsilon$ above can be quantified in terms of the machine precision, which in our case is $\varepsilon_{\mathrm{machine}} = \mathcal{O}(10^{-16})$. The heuristics is that if $\varepsilon_{\mathrm{machine}} = \mathcal{O}(10^{q})$ and $\kappa(\Psi) = \mathcal{O}(10^{q^{\prime}})$, the computed solution has about $q - q^{\prime}$ correct decimal digits. Hence, we say $\Psi$ is ill-conditioned with respect to the machine precision whenever $\varepsilon_{\mathrm{machine}} \kappa(\Psi)$
is large. For instance, if the condition number grows beyond $10^{13}$, the solution can have at best digits around $0.001$. The other quantity is the closeness of fit. If the closeness of fit is large, the relative error might scale with the second term in \cref{eq:rel_error_kappa}, i.e., $\kappa(\Psi)^2$. {Consequently, an ill-conditioned $\Psi$ might lead to large numerical inaccuracies. The Tikhonov regularization ($\beta > 0$) is an option to avoid the ill-conditioning, where $\kappa(\Psi)$ is replaced by the regularized condition number} $\kappa_{\beta}(\Psi) = \frac{\sigma_1}{\sqrt{\beta}}$ \cite[Theorem 5.1.1]{Hansen_1998}. By increasing $\beta$, the condition number becomes smaller, and thus the regularized solutions are less sensitive to perturbations. 

{
\subsection{Numerical algorithms for the regularized least squares}
\label{sec:num_algs}

Different numerical algorithms are available to compute the regularized least squares solution. For ill-conditioned problems, the computed solutions can vary drastically, since the numerical stability of these algorithms depends differently on the condition number. Although this issue is well-known in numerical analysis \cite{GoluVanl96,trefethen1997numerical,Higham_2002}, this has not been examined in the context of reservoir computing. To compare the impact of different solvers, we adopt three approaches used in the reservoir computing literature to solve the regularized least squares problem. All algorithms are implemented in \texttt{python} --- \texttt{scipy} \cite{2020SciPy-NMeth} package.}:
\begin{itemize}
    \item[] \textit{Cholesky decomposition:} The first method is based on solving the least squares via the regularized normal equation for the unknown $\us$:
    \begin{align}\label{eq:normal_equation}
        (\Psi^{\top} \Psi + \beta \mathbf{1}_m) \us = \Psi^{\top} \y_i, \quad i = 1,\dots, d,
    \end{align}
    which is computed numerically by Cholesky factorization in \texttt{scipy.linalg.solve}, assuming that the matrix is \emph{positive definite}. The relative error for solving linear equations is slightly different from \cref{eq:rel_error_kappa}: it scales quadratically with the condition number $\kappa(\Psi)$, because of the factor $\Psi^{\top} \Psi$, and does not depend on $\theta$ (see \cite{GoluVanl96,trefethen1997numerical,Higham_2002} for detailed exposition). For $\beta > 0$, we expect to scale as $\sigma_1^2/\beta$. 
    \item[] \textit{Singular Value Decomposition (SVD)}: The second method is written in terms SVD of $\Psi = \U \Sigma \V^{\top}$, which is computed via \texttt{scipy.linalg.svd}, with $\Sigma = \mathrm{diag}(\sigma_1, \dots, \sigma_{m})$,
    \begin{align}\label{eq:regularized_solution}
        \w_i = \Psi_{\beta}^{\dagger} \y_i,
    \end{align}
    where $\Psi_{\beta}^{\dagger} := \V \Sigma_{\beta} \U^{\top}$ is the regularized pseudo-inverse with the regularized singular values:
    \begin{align*}
        \Sigma_{\beta} = \mathrm{diag} \Big(\frac{\sigma_1}{\sigma_1^2 + \beta}, \dots, \frac{\sigma_{m}}{\sigma_{m}^2 + \beta} \Big).
    \end{align*}
    Although this method can be computationally expensive, it is a viable method to solve numerically the Tikhonov regularization \cite{Hansen_1998}. Solving the least square problem with SVD is known to be stable \cite[Theorem 19.4]{trefethen1997numerical}\footnote{The built-in implementation of SVD is based on a QR decomposition \cite{lapacklugnode53}. Solving least square problems based on QR decomposition satisfies \cref{eq:rel_error_kappa}.}. 
    \item[] \textit{LU decomposition}: The third method corresponds to solving \cref{eq:solution_Tikhonov} as it is mathematically written, where the inverse is computed numerically using LU decomposition via \texttt{scipy.linalg.inv}. This is similar to Choleskey decomposition, scaling quadratically with $\kappa(\Psi)$ for $\beta = 0$ and as $\sigma_1^2/\beta$ for $\beta > 0$.
\end{itemize}
Other commonly used numerical algorithms include \texttt{scipy.linalg.pinv} and \texttt{scipy.linalg.lstsq}, but those inherently add a built-in regularizer using the definition of $\varepsilon-$rank$(\Psi)$, where singular values are considered zero \cite[Theorem 2.5.3]{GoluVanl96} whenever they are upper bounded by $\sigma_1 \times \varepsilon_{\mathrm{machine}}$.

\begin{figure*}[t!]
\centering
\includegraphics[width=1.0\linewidth]{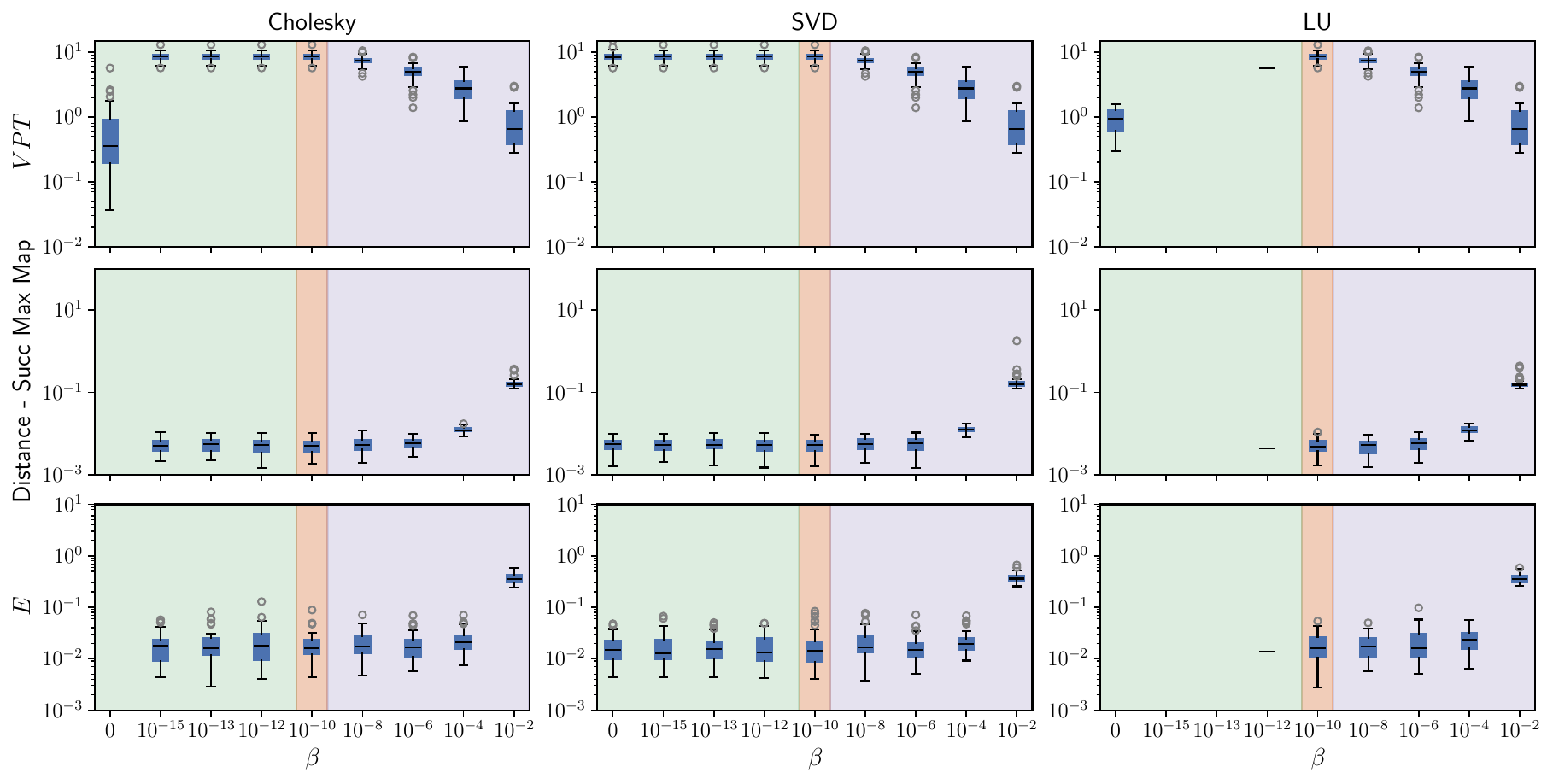}
\caption{{\textbf{Regularization also reduces the difference among numerical algorithms during NGRC testing.} Comparison between three different numerical algorithms: Cholesky, SVD, and LU as the regularizer parameter $\beta$ is increased under the testing phase. From top to bottom, panels show box plots with respect to bounded models over 50 different initial conditions:} valid prediction time (VPT), the distance between the induced successive local maxima maps, and the error $E$ between the power spectrum density, respectively. The colored hashed areas correspond to: absent/small (green), large (purple) regularization, and level of regularization in which all algorithms perform similarly within statistical confidence (orange). {Blank spaces correspond to unbounded NGRC models over all initial conditions.}. The parameters are $h = 0.01$, delay dimension $k = 2$, time lag $\tau = 1$, the maximum degree $p = 2$, and $\Ntrain = 5000$.}
\label{fig:algorithm_dependency}
 \end{figure*}

{
\subsection{Numerical algorithms in the ill-conditioned regime}
\label{sec:numerical_algo_behavior}

The Lorenz system generated by the explicit Euler method provides a clear example of the interplay between hyperparameters and numerical accuracy during NGRC training. By changing a single hyperparameter, in this case the delay dimension $k$, the NGRC model can transition from a well-conditioned regime to an extremely ill-conditioned regime, with the condition number scaling as the inverse of $\varepsilon_{\mathrm{machine}}$. In this ill-conditioned regime, the solutions obtained by different numerical algorithms described in \cref{sec:num_algs} can differ by orders of magnitude.

First, consider again the case $k = 1$, using the parameters from \cref{fig:NGRC_k_1_dt_001}. The condition number is small, $\kappa(\Psi) = 149.24$, and the closeness of fit for each Lorenz coordinate across the three algorithms is shown in \cref{table:closeness_of_fit}. Because the NGRC model is polynomial and fits a polynomial vector field, the closeness of fit is consistently of order $10^{-14}$, indicating a high fitting accuracy. Consequently, the maximum pairwise difference between solutions as defined in 
\begin{align}\label{eq:pairwise_diff}
\begin{split}
    \Delta = \max_{i = x, y, z} \max \Big\{ \|\w_i^{\mathrm{CHO}} - \w_i^{\mathrm{SVD}}\|, \|\w_i^{\mathrm{CHO}} - \w_i^{\mathrm{LU}}\|, \\\|\w_i^{\mathrm{SVD}} - \w_i^{\mathrm{LU}}\|\Big\},
\end{split}
\end{align}
is also on the order of $10^{-14}$, consistent with the error bounds predicted by \cref{eq:rel_error_kappa}.

\begin{table}[ht]
\centering
\begin{tabular}{|c|c|c|c|}
\hline 
Algorithm & $\theta_x$ & $\theta_y$ & $\theta_z$ \\
\hline \hline
Cholesky  & $1.51579 \times 10^{-14}$ & $1.39049\times 10^{-14}$ & $9.38594\times 10^{-15}$ \\  \hline
SVD & $2.95307\times 10^{-15}$ & $2.77759 \times 10^{-15}$ & $3.37643\times 10^{-15}$ \\ \hline
LU & $1.85695\times 10^{-14}$ & $2.28425\times 10^{-14}$ & $1.89852\times 10^{-14}$ \\\hline
\end{tabular}
\caption{Closeness of fit for each coordinate of the Lorenz system. The parameters are the same as those used for \cref{fig:NGRC_k_1_dt_001}.}
\label{table:closeness_of_fit}
\end{table}

Nevertheless, the situation drastically changes for $k = 2$. For the same initial condition used before, this hyperparameter choice, common in NGRC experiments \cite{Gauthier2021,Gauthier_2022_basins,Zhang_2023}, drives the model into an ill-conditioned regime, with $\kappa(\Psi) = 7.72 \times 10^{16}$ (essentially the inverse of machine precision). In this setting, the numerical algorithms diverge as shown by the closeness of fit for each algorithm in \cref{table:closeness_of_fit_k_2}. While the SVD algorithm still produces solutions with high fitting accuracy, the other algorithms perform poorly. Specifically, the Cholesky algorithm fails, as the matrix becomes numerically singular. The LU algorithm produces invalid solutions because the $\arcsin$ function argument was larger than 1. 

\begin{table}[ht]
\centering
\begin{tabular}{|c|c|c|c|}
\hline 
Algorithm & $\theta_x$ & $\theta_y$ & $\theta_z$ \\
\hline \hline
Cholesky  & $-$ & $-$ & $-$ \\  \hline
SVD & $4.11460\times 10^{-15}$ & $3.05268 \times 10^{-15}$ & $1.14009\times 10^{-14}$ \\ \hline
LU & $-$ & $0.70344$ & $-$ \\\hline
\end{tabular}
\caption{Closeness of fit for each coordinate of the Lorenz system. The symbol $-$ indicates that the value could not be computed. The parameters are delay dimension $k = 2$, and the rest of the parameters are the same as those used for \cref{fig:NGRC_k_1_dt_001}.}
\label{table:closeness_of_fit_k_2}
\end{table}

To reduce these issues, the regularizer parameter is varied as shown in \cref{fig:trade_off}. For better visualization, we introduce the maximum closeness of fit over all Lorenz coordinates as $\theta_{\max} = \max \{\theta_x, \theta_y, \theta_z\}$. The top panel shows that increasing $\beta$ gradually reduces the condition number to $\mathcal{O}(10^8)$, but also the fitting accuracy, inherent to the regularization trade-off. Small values of $\beta$ restore the existence of Cholesky solutions and bring their fitting accuracy closer to that of SVD solutions, although LU remains less accurate. By increasing $\beta$, all three algorithms have fitting performance that coincide at $\beta = 10^{-10}$, and converge to similar solutions, as shown in the bottom panel in \cref{fig:trade_off}. 

\begin{figure}
\centering
\includegraphics[width=1.0\linewidth]{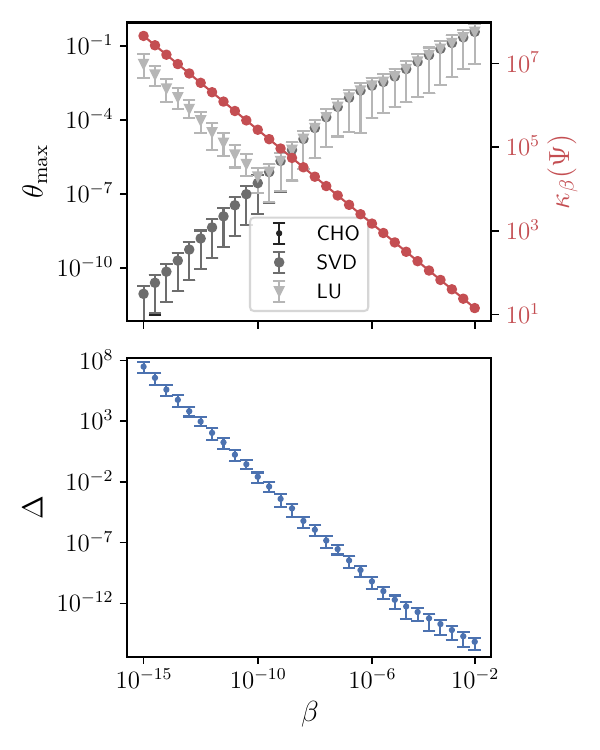}
\caption{\textbf{Regularization reduces the difference among numerical algorithms during NGRC training.} The top panel shows the maximum closeness of fit for Cholesky (dark gray), SVD (gray), and LU (light gray) solvers plotted together with the regularized condition number of $\Psi$ (red) with respect to the regularizer parameter $\beta$. All numerical algorithms attain similar fitting accuracy at $\beta = 10^{-10}$. While the fitting performance gets worse as the regularizer parameter is increased, the bottom panel shows that the maximum pairwise difference among the solutions decays monotonically. The dots are the median, and bars are $75\%$ and $25\%$ quantiles over 50 different initial conditions. The parameters are $h = 0.01$, delay dimension $k = 2$, time lag $\tau = 1$, the maximum degree $p = 2$, and $\Ntrain = 5000$.}
\label{fig:trade_off}
 \end{figure}

This trade-off between fitting accuracy and numerical stability extends to the testing phase. We select a few regularizers used during training and contrast the results of \cref{fig:trade_off} with the NGRC's resulting dynamics over multiple time steps, see \cref{fig:algorithm_dependency}}. The NGRC performance during testing is quantified by topological and statistical metrics, as described in detail in \cref{sec:metrics}: valid prediction time (VPT) \cite{Vlachas_2020} quantifies for how long, in Lyapunov times, the NGRC trajectory is close to the original one; the distance between the induced successive local maxima map; and divergence error $E$ between the power spectrum densities estimated from the trajectories. Note that the columns of $\Psi$ are not normalized, as normalization can lead to different solutions \cite{GoluVanl96}. {Whenever the algorithm failed to find a solution, $\w_i(\beta) = 0$. Moreover, if the trajectory of the NGRC during testing was unbounded, the metric value is set to $-1$. }

Starting from the right {in \cref{fig:algorithm_dependency}}, the purple region corresponds to the regime of large $\beta$. The VPT decreases monotonically, where all three algorithms obtain a solution whose performance drops because the regularization is too strong. The distance between successive maps and the divergence error remain statistically indistinguishable across initial conditions, until they deteriorate due to excessive regularization. The orange region corresponds to the regularizer value {$\beta = 10^{-10}$} that all algorithms perform equally over all metrics, {and coincides with the value at which all algorithms have similar fitting accuracy shown in \cref{fig:trade_off}.} This regime corresponds to sufficient regularization for the best performance in all three metrics.  As opposed to a smaller $\beta$, {in the green region}, where the three algorithms yield different NGRC performances. Cholesky and LU {algorithms} exhibit discontinuous behavior as $\beta$ changes, whereas SVD varies smoothly. SVD is robust against the ill-conditioning of $\Psi$. Surprisingly, at $\beta = 0$, its NGRC performance remains unaltered. {Thus, for this example, regularization is unnecessary for the SVD algorithm.} 

Before proceeding, {another benefit of studying the Lorenz example is that it sheds light on the relation between a large condition number and the autonomous NGRC dynamics. For a small time step $h$,} when training NGRC, the task reduces to finding the governing equations of a dynamical system directly from time series data \cite{Crutchfield_1987,Brown_1994_PLA,Brown_1994_PRE}. This {gives} an intuitive perspective into the autonomous NGRC dynamics during testing. Even small modeling errors --- especially for terms that are absent in the true system --- can cause instability when fitting an ODE from data, as already identified in \cite{Yao_2007}. For example, consider we attempt to learn $\dot{x} = - x,\dot{y} = 0$ but instead recover $\dot{x} = - x, \dot{y} = \varepsilon y$. Although $\varepsilon$ may be small, the trajectory of the learned system diverges due to exponential growth in $y$. Similarly, if $\Psi$ is ill-conditioned, the estimated readout weights may differ substantially from the true ones, leading to unstable NGRC dynamics \cite{Zhang_2025}.

\section{Problem statement}
\label{sec:problem_statement}
{The Lorenz example generated by the explicit Euler method demonstrates that the condition number $\kappa(\Psi)$ determines the accuracy of the NGRC model during training. In the ill-conditioned regime, the NGRC performance becomes highly sensitive to the numerical algorithm employed to solve the minimization problem in \cref{eq:w_out_training}. While regularization mitigates the effect of ill-conditioning, it introduces an extra hyperparameter, whose selection is non-trivial and has been studied extensively \cite{Hansen_1998}. In particular, for forecasting tasks, the common approach is to train the model to predict one step forward, and then select the regularizer parameter that ensures accurate prediction over multiple time steps ahead. Consequently, finding the appropriate $\beta$ inevitably runs into extra computational cost --- a cost that, for certain hyperparameter choices, could potentially be avoided. As shown using the SVD algorithm in the above example, NGRC can achieve accurate forecasts without any regularization.

This raises a fundamental question: can appropriate hyperparameter choices eliminate the need for regularization? To address this, we thoroughly investigate how the condition number varies with the hyperparameters: the delay dimension $k$, the time lag $\tau$, and the length of the training data $\Ntrain$.} Our focus here is to investigate under which conditions $\Psi$ becomes ill-conditioned. We observe that $\Psi$ possesses rich internal structure: for $k > 1$, it is simultaneously Vandermonde-like and Hankel-like because each entry corresponds to a monomial evaluated along a time-delayed trajectory of the original dynamics. As we will discuss, analyzing the conditioning of $\Psi$ is non-trivial and requires combining tools from numerical linear algebra with ergodic properties of dynamical systems. {Then, we utilize this mapping as a guide for better choices of hyperparameters, enabling accurate NGRC models without the need for regularization.}

\section{The structure of $\Psi$ characterizes its conditioning}
\label{sec:condition_number_characterization}

{By studying time series of chaotic systems,} we examine the matrix $\Psi$ in detail, observing that its structure resembles Vandermonde-like and Hankel-like matrices --- objects that frequently arise in interpolation, signal processing, and numerical linear algebra \cite{olshevsky2001structured, olshevsky2001structuredII}. We numerically demonstrate that $\Psi$ can become ill-conditioned under various parameter settings. The underlying causes are analyzed separately: the influence of the maximum polynomial degree, the interplay between time lags and delay dimension, and the length of training data.

{In all numerical experiments of this section, the numerical integrator used is the explicit Euler method, discarding the initial transient time of $10000$ time steps. Moreover, } since we are interested in the order of magnitude rather than the exact values of the condition number, we normalize $\Psi$ such that its columns have unit norm: for each column $\us_i$ of $\Psi$, the new columns are of the form $\frac{\us_i}{\|\us_i\|}$. This normalization is called column weighting \cite{GoluVanl96}, which realizes a lower bound for the condition number of the original matrix \cite{vanderSluis1969,Beckermann2000}. {The normalized matrix is denoted $\hat{\Psi}$.}

\subsection{Condition number grows exponentially with respect to $p$}
\label{sec:condition_number_exp_poly_deg}

\begin{figure}[t!]
\centering
\includegraphics[width=1.0\linewidth]{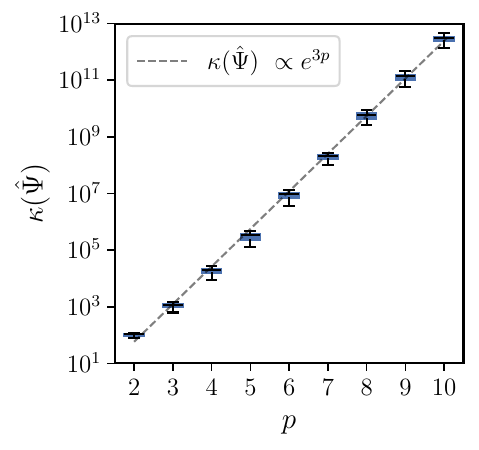}
\caption{\textbf{Exponential growth {with respect to the} maximum degree.} Box plot of the condition number $\kappa(\hat{\Psi})$ over 25 distinct initial conditions for increasing maximum degree $p$. The exponential growth is confirmed by the exponential curve (black dashed line) plotted for reference. The numerical integrator is the explicit forward Euler method with step size $h = 0.01$, and the number of training data points $\Ntrain = 5000$.}
\label{fig:exponential_growth_p}
 \end{figure}

The first numerical experiment increases the NGRC model's maximum degree and evaluates the condition number $\kappa(\hat{\Psi})$ over different initial conditions. \cref{fig:exponential_growth_p} shows that $\kappa(\hat{\Psi})$ increases exponentially with respect to $p$, with a growth rate approximately scaling as $e^{3p}$, as confirmed by the linear fit on the logarithmic scale. This suggests that, for systems requiring NGRC models with high-degree polynomials, the associated matrix $\Psi$ is unavoidably ill-conditioned.

The $\Psi$ structure can be related to a well-known matrix in numerical analysis and polynomial interpolation \cite{Higham_2002}. Note that $\Psi$ evaluates multivariate polynomials along the time series, yielding a Vandermonde-like structure \cite{Kuian_2019}. For example, for $k = 1$, $d = 2$ and $p = 2$:
\begin{align*}
   {\footnotesize
    \Psi = \begin{pmatrix}
        1 & x_{0, 1} & x_{0, 1}^2 & x_{0, 2} & x_{0, 2}^2 & x_{0, 1}x_{0, 2}\\
        1 & x_{1, 1} & x_{1, 1}^2 & x_{1, 2} & x_{1, 2}^2 & x_{1, 1}x_{1, 2}\\
        \vdots & \vdots & \vdots & \ddots & \vdots & \vdots\\
        1 & x_{\Ntrain - 1, 1} & x_{\Ntrain - 1, 1}^2 & x_{\Ntrain - 1, 2} & x_{\Ntrain - 1, 2}^2 & x_{\Ntrain - 1, 1}x_{\Ntrain - 1, 2}
    \end{pmatrix}.
    }
\end{align*}
There is a large body of evidence that Vandermonde matrices tend to be badly ill-conditioned \cite{Pan_2016_ill_conditioned_Vandermonde}. For the univariate case, the ill-conditioning is a consequence of the monomials being a poor basis for the polynomials on the real line. Monomials become increasingly highly correlated as the degree becomes larger over a finite interval. More specifically, it has been shown that the condition number of Vandermonde matrices for the univariate case has exponential growth with respect to $p$; see \cite{Gautschi1987,Kuian_2019}. In particular, the growth rate is at least $\big(\frac{2}{p + 1}\big)^{1/2} (1 + \sqrt{2})^{p - 1}$ for any choice of real positive or symmetrically distributed points (nodes) \cite{Beckermann2000}; for other lower bounds estimates, see \cite{Higham_2002}. Although precise results for the multivariate case (as relevant for NGRC) are lacking, our numerical observations are consistent with the exponential growth reported in the univariate case.

\subsection{The interplay between time lags and delay dimension}
\label{sec:interplay_time_lag_delay_dimension}

\begin{figure*}[t]
\centering
\includegraphics[width=0.8\linewidth]{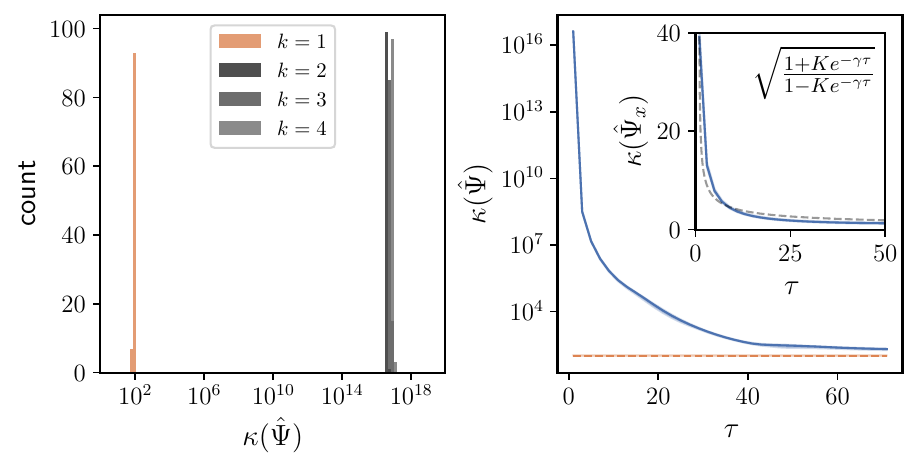}
\caption{\textbf{Time lag makes $\Psi$ better conditioned.} The left panel corresponds to the histogram of the condition number $\kappa(\hat{\Psi})$ over 100 distinct initial conditions for different combinations of NGRC models. {Bars corresponding to $k \geq 2$ overlap and form a cluster at $10^{16}$ over the different initial conditions, indicating that the matrix becomes ill-conditioned in the presence of delayed coordinates}. The right panel shows that the condition number monotonically decreases for larger time lag $\tau$. The solid line corresponds to the median, and the shaded areas correspond to $25\%$ and $75\%$ quantiles over 25 different initial conditions. The orange dashed line corresponds to the NGRC model with $k=1$, as shown in the left panel. The inset corresponds to the condition number of the submatrix $\hat{\Psi}_x$ for increasing $\tau$, whose decaying behavior is captured by the expression written for positive constant $K$ and exponent $\gamma$; see \cref{ex:one_dimensional} for details. The numerical integrator is the explicit forward Euler method with step size $h = 0.01$, maximum degree $p = 2$, and the number of data points $\Ntrain = 5000$.}
\label{fig:ill_conditioned_delay}
 \end{figure*}

Different from the maximum degree $p$, which is a known issue in any polynomial interpolation using a monomial basis \cite{Higham_2002}, the effect of adding time-delayed coordinates and time lag is more specific to time-delayed models, such as the NGRC model. The numerical observation is that $\Psi$ is ill-conditioned for small $\tau$. The left panel of \cref{fig:ill_conditioned_delay} shows the histogram of $\kappa(\hat{\Psi})$ over $100$ different initial conditions for different $k$. The case $k = 1$ (in orange) corresponds to the NGRC model akin to the explicit forward Euler method --- which will be the reference value (order $10^2$). As the delay dimension $k$ increases, the condition number abruptly shifts to $10^{16}$, indicating that the matrix has become rank deficient. We follow a similar argument \cite{Zhang_2025} to pinpoint the specific reason:

\begin{result}
    For small values of time lag $\tau$ and step size $h$, $\Psi$ is rank deficient for any delay dimension $k > 1$.
\end{result}

\begin{proof}
    Consider the case $\tau = 1$. For $k > 1$, there exists a subset of monomials in $\mathcal{P}_{p}^{kd}$ that evaluates the trajectory at consecutive time points, and their corresponding columns in $\Psi$ are linearly dependent on each other. This implies that the minimum singular value is close to zero, and consequently, $\Psi$ is rank deficient. For any univariate monomial in $\mathcal{P}_p^{kd}$, the argument is the following: choose two distinct columns of $\Psi$ corresponding to the same monomial evaluated at two consecutive time points. Let us denote such monomial as $\phi:\mathbb{R}^d \to \mathbb{R}$. By the explicit forward Euler method in \cref{eq:Euler_method} and $\phi$ being continuous differentiable function, the Mean-Value theorem implies that
\begin{align}\label{eq:Mean_value_theorem_basis_function}
    \phi(\x_{n + 1}) = \phi(\x_{n}) + h \Big(\int_{0}^1 \nabla \phi(\x_{n} + s \F(\x_{n})) ds\Big) \cdot \F(\x_{n}),
\end{align}
where $\cdot$ corresponds to the usual inner product in $\mathbb{R}^d$. Two columns of $\Psi$ corresponding to $\phi$ evaluated along the time series for consecutive time points are given as
\begin{align}\label{eq:column_vectors}
    \us_1 = \frac{1}{\sqrt{\Ntrain}} \begin{pmatrix}
        \phi(\x_{0})\\
        \phi(\x_{1}) \\
        \vdots \\
        \phi(\x_{\Ntrain - 1})
    \end{pmatrix} ~ \text{and} ~  \us_2 = \frac{1}{\sqrt{\Ntrain}} \begin{pmatrix}
        \phi(\x_{1})\\
        \phi(\x_{2}) \\
        \vdots \\
        \phi(\x_{\Ntrain})
    \end{pmatrix}.
\end{align}
If any two columns of $\Psi$ are equal, it implies that $\Psi$ is singular, i.e., the minimum singular value $\sigma_m(\Psi) = 0$. Consequently, by continuity of the singular values in terms of the entries of $\Psi$, $\us_2 = \us_1 + \mathcal{O}(h)$ with small $h$ implies that $\sigma_m(\Psi)$ is close to zero. 
\end{proof}

A similar argument can be made for $\hat{\Psi}$, and other forward numerical integration schemes applied to the original differential equations. In agreement with the left panel of \cref{fig:ill_conditioned_delay}, the decreasing order of singular values of $\Psi$ has a clear gap, confirming that the matrix is rank deficient - results not shown. 

We observe that the submatrices formed by evaluating the same monomial at different time-delays resemble a Hankel-like structure, which has already been analyzed in the context of data-driven dynamical systems in Dynamic mode decomposition \cite{Arbabi_2017_hankel_DMD}:
\begin{align*}
    \frac{1}{\sqrt{\Ntrain}} \begin{pmatrix}
        \phi(\x_{0}) & \phi(\x_{1}) & \dots & \phi(\x_{(k - 1)\tau})\\
        \phi(\x_{1}) & \phi(\x_{1 + \tau})& \dots & \phi(\x_{1 + (k -1)\tau})\\
        \vdots & \vdots & \ddots & \vdots \\ 
        \phi(\x_{\Ntrain - 1}) & \phi(\x_{\Ntrain - 1 + \tau}) & \dots & \phi(\x_{\Ntrain - 1 + (k - 1)\tau})
    \end{pmatrix}. 
\end{align*}
We utilize this structure to gain further insights on the influence of the time lag $\tau$ on the condition number, as we detail below.

\subsubsection{Increasing the time lag improves the conditioning of $\Psi$} Although the exponential growth with respect to the maximum degree is inevitable, the ill-conditioning due to small $\tau$ vanishes for large time lags. A fixed maximum degree and increasing the time lag make $\Psi$ better conditioned; see the right panel of \cref{fig:ill_conditioned_delay}. It shows that for increasing the time lag, $\kappa(\hat{\Psi})$  decreases and asymptotically approaches the reference value (orange line), which corresponds to the condition number of the NGRC model with $k = 1$ and $p = 2$ as used in the left panel of \cref{fig:ill_conditioned_delay}. Increasing the time lag eliminates the spurious linear dependence between columns evaluating consecutive time points, which has also been observed in magnetic pendulum dynamics \cite{Zhang_2025}. 

The monotonic decay with respect to the time lag can be related to the statistical properties of the original dynamics:  computing the condition number requires computing the singular values of $\hat{\Psi}$, which are defined to be the square root of eigenvalues of $\hat{\Psi}^{\top} \hat{\Psi}$. First note that $\Psi^{\top}\Psi$ corresponds to the Euclidean inner product of columns $\us_i$ of $\Psi$, as in \cref{eq:column_vectors}, where each entry is given by
\begin{align}\label{eq:birkhoff_psi}
\begin{split}
    (\Psi^{\top} \Psi)_{ij} = \langle \us_i, \us_j \rangle &= \frac{1}{\Ntrain} \sum_{n = 0}^{\Ntrain - 1} \psi_i(\X_n) \psi_j(\X_n) \\
    &= \frac{1}{\Ntrain} \sum_{n = 0}^{\Ntrain - 1} (\psi_i \psi_j)\circ \g_{k, \tau}(\f^{n}(\x_0))\\
    & =: \frac{1}{\Ntrain} S_{\Ntrain}((\psi_i \psi_j) \circ \g_{k, \tau})(\x_0),
\end{split}
\end{align}
which corresponds to a Birkhoff average of the observable $(\psi_i \psi_j) \circ \g_{k, \tau}:M \to \mathbb{R}$. Then, the normalized version has the form $(\hat{\Psi}^{\top} \hat{\Psi})_{ij} = \langle \vs_i, \vs_j \rangle $, where $\vs_i := \frac{\us_i}{\|\us_i\|}$.

Since we are dealing with a chaotic system, it naturally calls for evoking its statistical properties. Let $\mu$ be an invariant ergodic measure preserved under $\f$. By the Birkhoff Ergodic theorem \cite{Viana_Oliveira_2016}: for any $\phi \in L^1(\mu)$:
\begin{align}\label{eq:birkhoff_thm}
    \lim_{N \to \infty} \sum_{n = 0}^{N - 1} \phi \circ \f^{n} (\x_0) = \int_{M} \phi d\mu, \quad \mu- \text{a.e.}
\end{align}
For sufficiently chaotic dynamics, which is commonly captured by the decay of correlations over time, consider the case when the original dynamics is \textit{exponential mixing}: $(\f, \mu)$ is exponential mixing, such that the correlation function for any $\phi, \varphi, (\phi \cdot \varphi) \in L^1(\mu)$ decays {exponentially for $\tau \geq 0$}:
\begin{align}\label{eq:exponential_mixing}
    \Big|\int \phi (\varphi \circ \f^{\tau}) d \mu - \int \phi d \mu \int \varphi d \mu \Big| \leq K(\phi, \varphi) e^{-\gamma \tau}.
\end{align}
{Such exponential decay of correlations \cite{viana1997stochastic} is a statistical property commonly found in chaotic systems, including the Lorenz system 63 \cite{Araújo2016}}. 

The Birkhoff Average theorem is valid for the observable $(\psi_i \psi_j) \circ \g_{k, \tau}$ if the pushforward probability measure $(\g_{k, \tau})_*\mu$ is an ergodic measure. This depends on the properties of the embedding map $\g_{k, \tau}$, which in turn depends on the delay dimension and time lag. Alternatively, since $\g_{k, \tau}$ evaluates monomial basis at different time, let us denote simply the monomial basis functions on the variables in $\mathbb{R}^d$  as $\phi_i: M \to \mathbb{R}$, i.e., $\phi_i \in \mathcal{P}_p^d$. Hence, the Birkhoff average in \cref{eq:birkhoff_psi} has a particular form {
\begin{align}\label{eq:birkhoff_average_S_train}
\begin{split}
   \frac{1}{\Ntrain} S_{\Ntrain}&((\psi_i \psi_j) \circ \g_{k, \tau})(\x_0) \\ 
   & = \frac{1}{\Ntrain} S_{\Ntrain}\big((\phi_i \circ \f^{q}) (\phi_j \circ \f^{q^{\prime}})\big) (\x_0),
\end{split}
\end{align}}
$i, j = 1, \dots, m$ and $q, q^{\prime} = 0, 1, \dots, \tau.$ By {\cref{eq:birkhoff_thm}}, {the Birkhoff average in \cref{eq:birkhoff_average_S_train} converges $\mu-$a.e. to}
{\small
\begin{align*}
\lim_{\Ntrain \to \infty}\frac{1}{\Ntrain} S_{\Ntrain}\Big((\phi_i \circ \f^{q}) (\phi_j \circ \f^{q^{\prime}})\Big){(\x_0)} &= \int_M (\phi_i \circ \f^{q}) (\phi_j \circ \f^{q^{\prime}}) d\mu \\ 
    &= \int_M \phi_i (\phi_j \circ \f^{(q^{\prime} - q)}) d\mu,
\end{align*}
}{where the last equality holds due to the invariance of $\mu$ under $\f$.} For an arbitrary $\f$ and the monomial basis $\mathcal{P}_p^d$, the right-hand side corresponds to the cross-correlation function for the observables $\phi_i$ and $\phi_j$ as in \cref{eq:exponential_mixing}. The exponential decay of cross-correlations between delayed observables implies that the spurious linear dependencies among columns of $\Psi$ diminish as $\tau$ increases, as shown by the right panel in \cref{fig:ill_conditioned_delay}. For the monomial basis, it is a hard problem to prove such a result. In contrast, we can examine a toy case to gain insight into this intuition, as shown in \cref{ex:one_dimensional}. 

\begin{example}[$x$-coordinate case]\label{ex:one_dimensional}
Consider the simplest case: $k = 2$, $d = 1$ and $p = 1$, corresponding to $\mathcal{P}_{1}^{2} = \{1, x_{n}, x_{n + \tau}\}$. We denote the submatrix formed only with these terms as $\hat{\Psi}_x$. Also, assume that the time series $\{x_n\}_{n\geq 0}$ has mean zero, which is approximately satisfied by the $x-$coordinate of the Lorenz system due to the system's symmetry. For this case, $\hat{\Psi}^{\top} \hat{\Psi}$ has the form
\begin{align*}
    \begin{pmatrix}
        1 & 0 & 0 \\
        0 & 1 & \langle \vs_2, \vs_3 \rangle \\
        0 & \langle \vs_3, \vs_2 \rangle & 1 \\
    \end{pmatrix},
\end{align*}
whose maximum and minimum eigenvalues are $1 + \langle \vs_2, \vs_3 \rangle$ and $1 - \langle \vs_2, \vs_3 \rangle$, respectively. Consequently, the condition number is given by
\begin{align}\label{eq:bound_cond_num}
\begin{split}
    \kappa(\hat{\Psi}) &= \sqrt{\frac{1 + \langle \vs_2, \vs_3 \rangle}{1 - \langle \vs_2, \vs_3 \rangle}} \\ 
    &\leq \sqrt{\frac{1 + K e^{- \gamma \tau}}{1 - K e^{- \gamma \tau}}} =: h_{K, \gamma}(\tau), 
\end{split}
\end{align}
for a positive constant $K$ and decay exponent $\gamma$, after applying the exponential decay of correlation bounds in \cref{eq:exponential_mixing}. This formula $h_{K, \gamma}(\tau)$ shows that: while the function can be unbounded for small $\tau$, the function decays to one for large $\tau$. The inset of the right panel in \cref{fig:ill_conditioned_delay} shows $\kappa(\hat{\Psi}_x)$ (solid line) for increasing time lag, which decays monotonically to one. The dashed line corresponds to the fitted curve (with $K = 1.01$ and $\gamma = 0.01$) by the function $h_{K, \gamma}(\tau)$ in \cref{eq:bound_cond_num} to the numerical value.  Interestingly, for $K = 1$, the function can be written as $h_{1, \gamma}(\tau) = \coth^{\frac{1}{2}}{\big(\frac{\gamma \tau}{2}\big)}$. 
\end{example}

We observe that the decay of $\kappa(\hat{\Psi})$ is slower than $\kappa(\hat{\Psi}_x)$, which is related to the presence of the additional monomials. For instance, other observables as the $z-$coordinate have slower correlation decay. A more detailed analysis of the full matrix will be left for future work. 

\begin{figure*}[t!]
\centering
\includegraphics[width=0.85\linewidth]{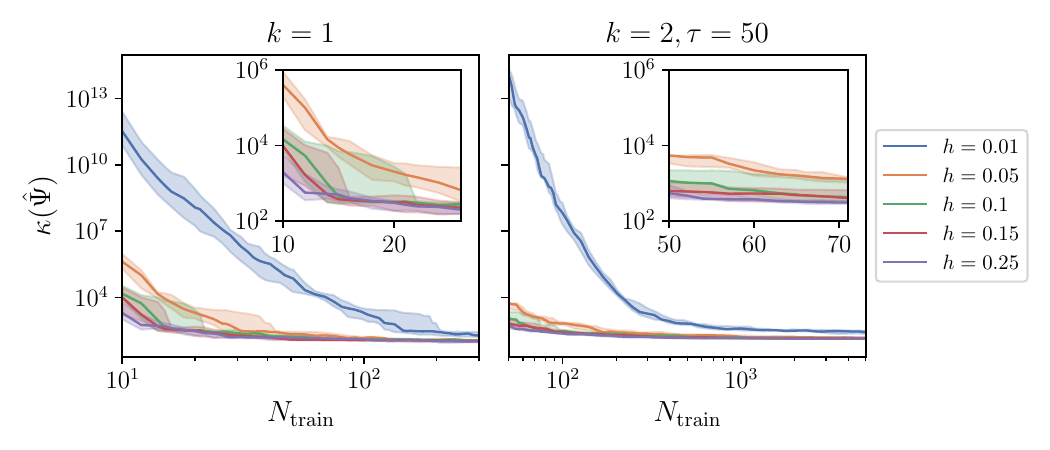}
\caption{\textbf{Condition number with respect to the length of training data.} Condition number of $\hat{\Psi}$ for increasing $\Ntrain$ for different time steps $h$. The left panel shows the results for the NGRC model with $k = 1$ ($m = 10$), and the right panel shows the model with $k = 2$, $\tau = 50$ ($m = 28$). In both cases, the NGRC model has a maximum degree $p = 2$. All curves for $h > 0.1$ overlap for all $\Ntrain$ tested. The solid line corresponds to the median, and the shaded areas correspond to $25\%$ and $75\%$ quantiles over 10 different initial conditions.}
\label{fig:NGRC_dependence_Ntrain}
 \end{figure*}
 
\subsection{Dependence on the length of training data}
\label{sec:dependence_lgth_training_data}
As we have seen, the statistical properties of the original dynamics give insights into the condition number of $\Psi$. This is not different from the length of training data $\Ntrain$. As aforementioned, to define the least squares problem during training, $\Psi$ should be tall and slim ($\Ntrain > m$); otherwise, the matrix has a non-trivial kernel and is rank deficient. To test the dependence on $\Ntrain$, \cref{fig:NGRC_dependence_Ntrain} shows the condition number of $\Psi$ for $\Ntrain > \binom{kd + p}{p}$. For moderate values of $\Ntrain$, which is not much larger than $\binom{kd + p}{p}$, the condition number is not small. We observe that the condition number decays monotonically toward a limiting value $10^{2}$, but not uniformly. Initially, the decay is much faster than compared to larger values of $\Ntrain$. In the left panel, the NGRC model corresponds to the reference model we have been using so far, $k = 1$ and $p = 2$, and the right panel corresponds to the NGRC model with more terms, $k = 2$, $\tau = 50$, and $p = 2$. 

Since we are dealing with an ordinary differential equation, the length of the training data is determined by the time step $h$. Both panels of \cref{fig:NGRC_dependence_Ntrain} show the dependence of $\kappa(\hat{\Psi})$ for different time steps. For this experiment, the numerical integration is performed using the explicit forward Euler method with time step $h = 0.01$, then the time series is subsampled accordingly to the larger time steps. Similarly to $\tau$, the convergence to the limiting condition number is faster as the time step $h$ increases. 

Notice that in both cases the condition number decays, but as expected, an NGRC model that contains more monomials requires more data points to reach the limit. In other words, the linear dependencies among columns can be long-lived as more monomials are used in the NGRC model. The convergence of $\kappa(\Psi)$ toward the limit and the difference between small and large $h$ can be explained due to the statistical properties of the Lorenz system, as we presented. The convergence follows from the Birkhoff averaging theorem in \cref{eq:birkhoff_thm}. For sufficiently long training data, the entries of $\hat{\Psi}^{\top} \Psi$ converge in a neighborhood of a limit value under different rates that depend on the pair of observables $(\psi_i \psi_j) \circ \g_{k, \tau}$. This is in agreement with a recent result in traditional reservoir computing, which has shown that the Tikhonov regularization is an $L^2(\mu)$ approximation of ergodic dynamics \cite{HART_2021}. 

Similarly, $h$ in this particular construction plays a similar role as $\tau$ before. Larger $h$ corresponds to skipping the time evaluation of the monomial along time, and the exponential decay of correlations results in a smaller condition number. The fact that the entries of the matrix converge to a limit motivates scaling the matrix by the factor $\frac{1}{\sqrt{\Ntrain}}$. This is also relevant for the regularizer parameter, as we detail {in the following remark.}

\begin{remark}[Scaled regularizer parameter $\beta$ by the length of training data $\Ntrain$] The scaling $\frac{1}{\sqrt{\Ntrain}}$ in the definition of $\Psi$ in \cref{eq:library_matrix} also scales $\beta$ with respect to the data length $\Ntrain$, which has been a common practice in statistics \cite{Gene_Golub_1979}, and more recently in reservoir computing \cite{Chepuri_2024,Zhang_2025}.
Let us denote $\Tilde{\Psi} = \frac{1}{\sqrt{\Ntrain}} \Psi$. It follows that when $\tilde{\Psi}$ is used, the coefficient $\w(\beta)$ in \cref{eq:solution_Tikhonov} can be mapped to another coefficient with an adjusted regularizer parameter $\Ntrain \beta$. Note that 
\begin{align*}
    \tilde{\w}(\beta) &= (\tilde{\Psi}^{\top} \tilde{\Psi} + \beta \mathbf{1}_m)^{-1}\tilde{\Psi}^{\top} \y_i \\
    &= \frac{1}{\sqrt{\Ntrain}} (\frac{1}{\Ntrain}{\Psi}^{\top} \Psi + \beta \mathbf{1}_m)^{-1}{\Psi}^{\top} \y_i \\
    &=\sqrt{\Ntrain} \w(\Ntrain \beta).
\end{align*}
\end{remark}
{
\section{Condition number guides better choices of hyperparameter}
\label{sec:condition_guides_hyperparameter}

The characterization of the condition number from \cref{sec:condition_number_characterization} can be used to guide better choices of hyperparameter for NGRC training. In this section, we evaluate NGRC in two application scenarios: (i) sparsely-sampled time series; and (ii) partial measurement, where only the $x$-coordinate of the Lorenz system is available. In both cases, we identify regions of the hyperparameter space where the NGRC model reproduces the original dynamics without any regularization, thereby avoiding the extra computational cost of tuning $\beta$. 

To test robustness against higher-accuracy numerical integrators, throughout this section, we utilize a fourth-order Runge–Kutta method with a fine time step of $0.001$, followed by subsampling by time step $h$. Also, the length of training data is sufficient to avoid ill-conditioning due to undersampling, as discussed in \cref{sec:dependence_lgth_training_data}. Additional numerical experiments with the electronic circuit model of the Double-Scroll introduced in \cite{Chang_1998_DoubleScroll_intro} generalize the findings to other dynamical systems, see Supplementary Material.

\subsection{Sparsely-sampled training data}
\label{sec:sparse_sampled}

\begin{figure*}[t!]
\centering
\includegraphics[width=1.0\linewidth]{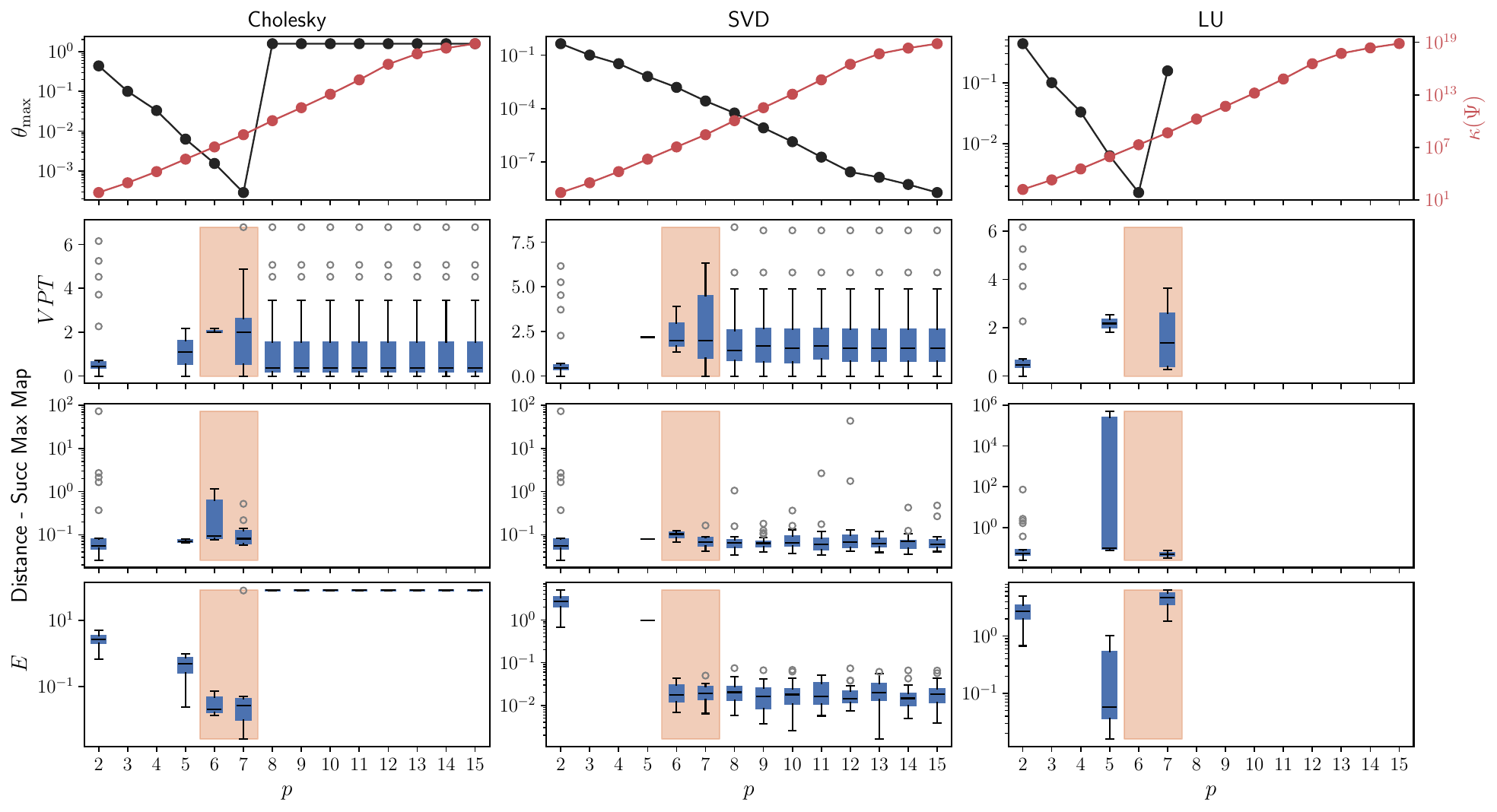}
\caption{\textbf{Sparsely-sampled time series requires an NGRC model 
 with a larger polynomial degree.}
The top row shows the maximum closeness of fit (black) and the condition number of $\Psi$ (red) across polynomial degrees, for the three numerical algorithms. Dots indicate the median over $25$ initial conditions. Subsequent rows show NGRC performance metrics as box plots over the different initial conditions and bounded models, similar to \cref{fig:algorithm_dependency}. Blank spaces correspond to unbounded NGRC models over all initial conditions. Training data were generated with a fourth-order Runge–Kutta and subsampled with step size $h = 0.1$. The parameters are delay dimension $k = 1$, time lag $\tau = 1$, $\Ntrain = 1000$, $\Ntest = 1000$, and $\beta = 0$.}
\label{fig:sparse_sampling}
 \end{figure*}

Consider the scenario where the Lorenz system is subsampled with a moderate time step $h = 0.1$. The input data exhibit more intricate dynamics than the original quadratic vector field. The effective dynamics result from a combination of the numerical approximation introduced by the numerical integrator and successive polynomial compositions between time samples. Consequently, capturing this behavior requires a higher-degree polynomial expansion so that the true dynamics lie within the span of $\mathcal{P}_p^{kd}$. In this context, we evaluate the NGRC performance as the maximum degree $p$ is increased, see \cref{fig:sparse_sampling}.

Increasing $p$ to match the original dynamics inevitably pushes the NGRC model toward an ill-conditioned regime, as analyzed in \cref{sec:condition_number_exp_poly_deg}. To examine this, we also test the consistency of the numerical algorithms' behavior discussed in \cref{sec:numerical_algo_behavior}. The top row of \cref{fig:sparse_sampling} displays the trade-off between the maximum closeness of fit $\theta_{\max}$ and the condition number as $p$ increases. The condition number grows exponentially, consistent with the same growth rate found for the explicit Euler method in \cref{sec:condition_number_exp_poly_deg}. Unlike increasing $\beta$ in \cref{fig:trade_off}, here the fitting accuracy improves with larger $p$, since a richer polynomial basis better represents the dynamics. Only the SVD algorithm captures this improvement smoothly, showing a monotonic increase in fitting accuracy. By contrast, the Cholesky and LU solutions degrade rapidly beyond $p > 7$ due to numerical instability.

To quantify the NGRC performance during the testing phase, we evaluate the valid prediction time, the distance between the induced successive local maxima maps, and the divergence error $E$. The main result is that, for certain values of $p$, NGRC can successfully reproduce the input data without any regularization. Moreover, examining how different numerical algorithms perform, Cholesky and SVD exhibit statistically similar performance within the orange shaded area, whereas the LU algorithm fails for all values of $p$. Interestingly, similar to the dependence on $\beta$, the SVD algorithm displays a smooth and stable dependence on $p$. It remains accurate across all tested values, with performance comparable to the orange shaded region even as $p$ increases. In contrast, Cholesky breaks down for $p > 7$, producing different dynamics from the original system. This demonstrates that the SVD algorithm is consistently more robust to the ill-conditioning of the feature matrix, making it a better choice when aiming for an accurate NGRC model without regularization. 
}

\subsection{Partial measurement from the Lorenz system}
\begin{figure*}[t!]
\centering
\includegraphics[width=0.9\linewidth]{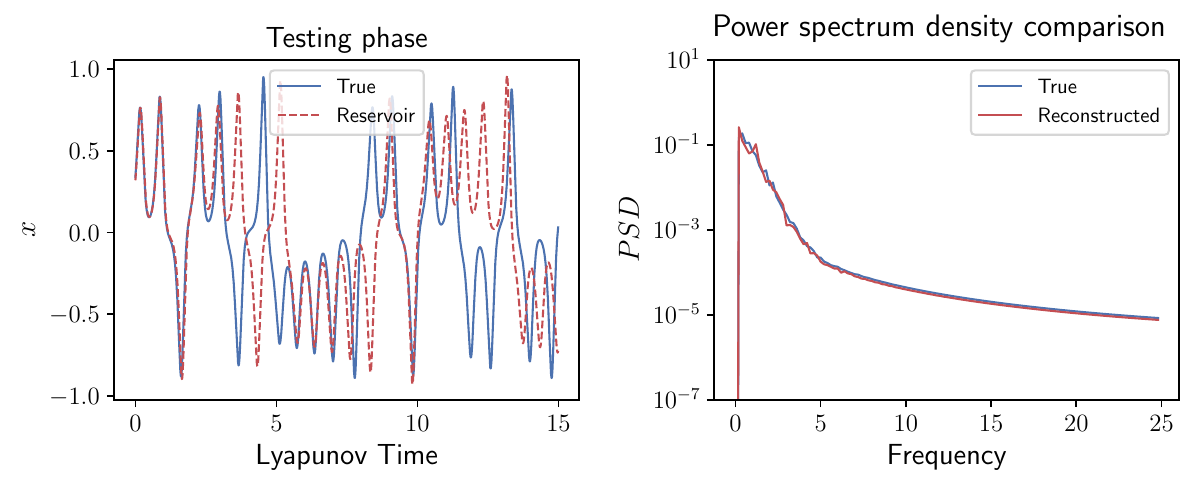}
\caption{\textbf{NGRC learns partial measurements of the Lorenz system.} The left panel shows the NGRC trajectory (red) compared to the original (blue). The right panel shows the comparison between the power spectrum densities. The numerical integration is the fourth-order Runge-Kutta with time step $h = 0.01$. The parameters are delay dimension $k = 3$, time lag $\tau = 15$, the maximum degree $p = 7$, $\Ntrain = 10000$, $\Ntest = 10000$ and regularizer parameter $\beta = 0$.}
\label{fig:x_coord_reconstr}
 \end{figure*}

\begin{figure*}[t!]
\centering
\includegraphics[width=1.0\linewidth]{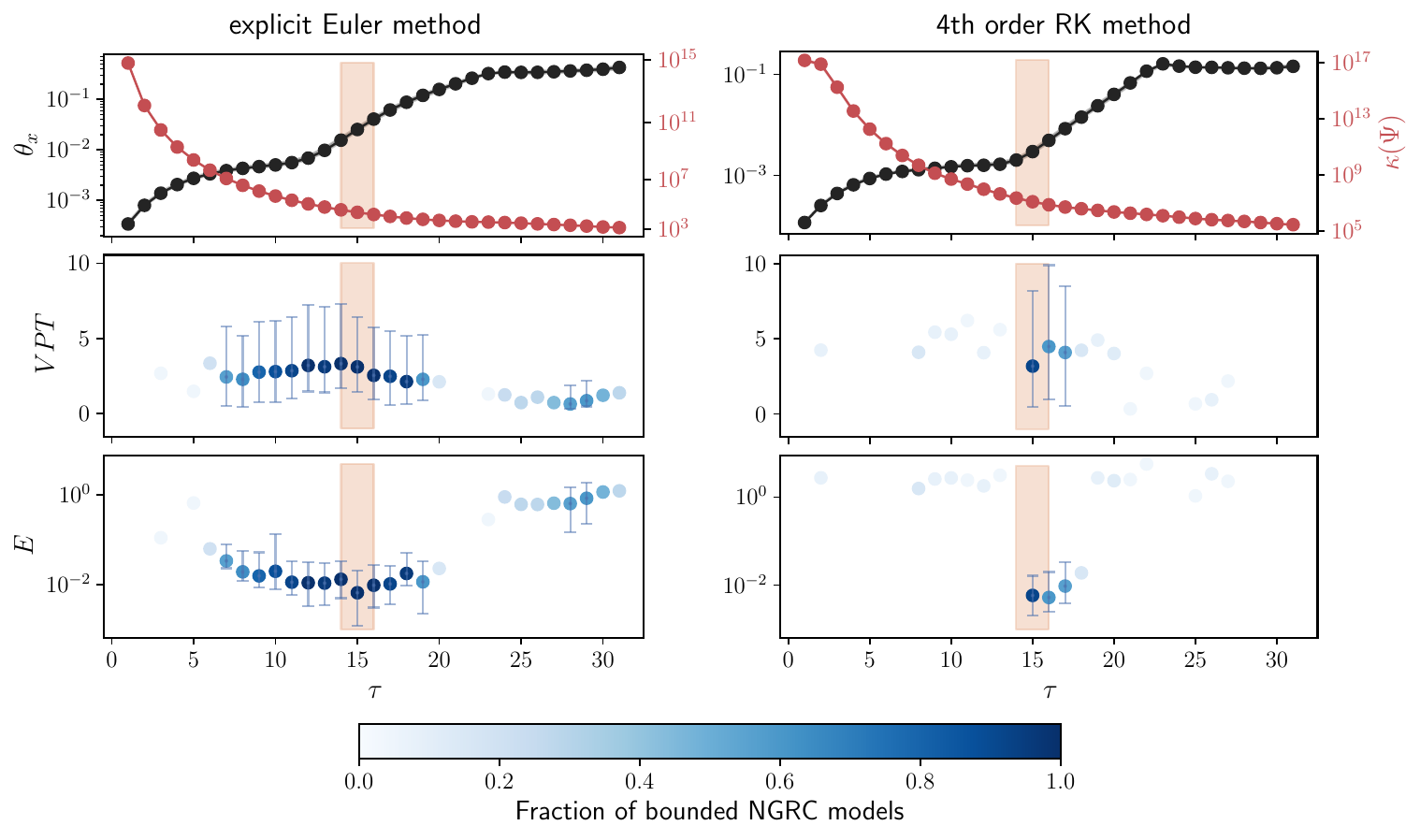}
\caption{\textbf{Methods used for embedding time series may inform NGRC modelling of partial measurements.} The top row displays the closeness of fit $\theta_x$ (black) and condition number of $\Psi$ (red) with respect to the increasing time lag $\tau$. The subsequent rows show the NGRC performance in both metrics, VPT and divergence error $E$. The orange band represents the $\tau = 15$, which is informed by the first minimum of mutual information. The blue colormap represents the fraction of bounded NGRC models over the $25$ initial conditions during testing. The dots correspond to the median, and error bars are $25\%$ (lower bound) and $75\%$ (upper bound) quantiles, which are shown only for NGRC models with a fraction larger than $0.5$. The maximum degree is $p = 5$ for the explicit Euler method, and $p = 7$ for the fourth-order Runge-Kutta method. The parameters are $h = 0.01$, delay dimension $k = 3$, $\Ntrain = 10000$, $\Ntest = 10000$ and regularizer parameter $\beta = 0$.}
\label{fig:x_coord_reconstr_time_lag}
 \end{figure*}

{
Consider a partial measurement of the Lorenz system, where only the $x$-coordinate time series $\{x_n\}_{n \geq 0}$ is available, as shown in the left panel of \cref{fig:x_coord_reconstr}. According to Takens' embedding theorem \cite{Packard_1980,Takens_1981}, the underlying dynamics in this case is an unknown function of the current and time-delayed coordinates, rather than the Euler discretization of a polynomial vector field. Consequently, this function does not necessarily lie in the span of $\mathcal{P}_p^{kd}$ and must instead be approximated.

Rather than performing a grid search over all hyperparameters, we analyze NGRC performance by varying only the time lag $\tau$, while keeping the delay dimension $k$ and maximum polynomial degree $p$ fixed. The delay dimension $k=3$ is chosen using the false nearest neighbors method \cite{Kantz_Schreiber_2003}, and the polynomial degree is set high enough to approximate the underlying dynamics. To demonstrate robustness, we test the forecasting task on data generated by two numerical integrators for $h = 0.01$: the explicit Euler method (as in earlier sections) and the fourth-order Runge–Kutta method, which accounts for subsampling. All simulations use the SVD algorithm, as it provides the most stable numerical algorithm. 

\cref{fig:x_coord_reconstr_time_lag} shows the closeness of fit $\theta_x$ and the condition number $\kappa(\Psi)$ as $\tau$ increases. For both numerical integrators, increasing $\tau$ leads to a monotonic increase in $\theta_x$ and a monotonic decrease in $\kappa(\Psi)$, consistent with our earlier discussion in \cref{sec:interplay_time_lag_delay_dimension}. In particular, for small $\tau$, the columns of the feature matrix are highly correlated, but as $\tau$ grows, they become increasingly uncorrelated. While this improves numerical stability, it reduces the NGRC model’s ability to capture the dynamics accurately. This might explain the plateau (visible in logarithmic scale) of $\theta_x$ for $\tau > 20$.

The NGRC performance is evaluated using the valid prediction time, the divergence error $E$ between the power spectrum densities, and the fraction of bounded models, defined as the percentage of initial conditions whose autonomous NGRC trajectory remains within the compact region $[-1,1]^3$. The key result is that, for both numerical integrators, NGRC can reproduce the partial measurement dynamics without any regularization, see an example trajectory and power spectrum for $\tau = 15$ in \cref{fig:x_coord_reconstr}. However, the dependence on $\tau$ differs between the two types of data. For the explicit Euler method, a broad range of $\tau$ values yields stable, bounded models with similar performance. In contrast, the Runge–Kutta data exhibit a narrower window of good performance, with only three $\tau$ values achieving a fraction larger than $0.5$ of bounded NGRC models. This difference likely arises from the additional complexity introduced by partial observation combined with subsampling in the Runge–Kutta data.

Interestingly, the optimal NGRC performance for both data types coincides with the time lag $\tau = 15$ corresponding to the first minimum of mutual information, a standard heuristic for selecting embedding time lags \cite{Fraser_1986,Kantz_Schreiber_2003}. This suggests that the method used for embedding time series can inform the choice of NGRC hyperparameters, reducing the need for exhaustive grid searches and accelerating training.

}

\section{Discussion and conclusions}
\label{sec:discussion_conclusions}

{
This work demonstrated that numerical analysis of NGRC training can guide better hyperparameter choices, often enabling training without any regularization. We have two main contributions. First,} by leveraging the Vandermonde- and Hankel-like structures of the feature matrix and the statistical properties of the {chaotic systems}, we elucidated the interplay between hyperparameters and feature matrix conditioning. Through numerical experiments, we demonstrated that the condition number of $\Psi$ can become prohibitively large: (i) high maximum degree $p$, (ii) short time delay $\tau$, and (iii) short length of training data $\Ntrain$. {Second, we showed that the choice of numerical algorithm critically impacts robustness to ill-conditioning. Across all scenarios tested, the LU algorithm was significantly less stable than the Cholesky and SVD algorithms. Moreover, there exist hyperparameter regions where SVD-based training achieves accurate forecasts without any regularization. This result may have gone unnoticed, as the common practice is to apply sufficiently strong regularization to ensure good NGRC performance, eliminating the large difference between the algorithms' solutions, as we have seen in \cref{sec:numerical_algo_behavior}.} This dependence on the numerical method suggests a practical step: prefer SVD-based training before deciding whether regularization is necessary. {In case SVD is} computationally expensive, alternatives for efficient Tikhonov regularization exist, including classical approaches such as Élden's method \cite{Elden_1977} and modern techniques for large-scale inverse problems \cite{Lampe_2012}.

{
Our analysis in \cref{sec:condition_guides_hyperparameter} further demonstrated that NGRC achieves superior forecasting performance in hyperparameter regions that balance numerical stability and fitting accuracy. In other words, large values of either the condition number or the closeness of fit tend to degrade performance. This observation suggests another practical heuristic: by monitoring $\kappa(\Psi)$ and the closeness of fit during training, one can readily identify hyperparameter regions unlikely to yield reliable long-term forecasts. When combined with the absence of regularization, this strategy may reduce training computational time.
}

One of the primary contributors to ill-conditioning is the use of high-degree polynomials. A natural strategy to mitigate this is to select a subset of monomials most relevant to the dynamics. However, identifying the optimal combination of terms is computationally challenging \cite{Davis1997}. Greedy algorithms, though suboptimal, have been successfully applied in reservoir computing \cite{Dutoit_2009}. More sophisticated approaches might impose prior structural information, such as sparsity or symmetry, aligning with recent developments in physics-informed machine learning and reservoir computing \cite{Chen2021,Matthew_2023,Gurevich_Golden_Reinbold_Grigoriev_2024,Barbosa_symmetry_awareness_2021}. Developing practical algorithms to construct NGRC models that preserve or approximate selected features of the original system remains an open direction. For complex dynamics, long polynomial expansions (large $p$) may {remain} unavoidable. In such cases, one must balance accuracy, interpretability, and computational cost. Recent alternatives address these challenges. \re{For interpretable models, local low-degree polynomial approximations based on NGRC have been proposed \cite{gauthier2025localityblendedgenerationreservoir}. For efficiency, departing from the equivalence of NGRC and polynomial kernel regression, recursive Volterra series expansions offer a delay-agnostic, lower-cost alternative to NGRC \cite{Grigoryeva_2025_infinite_dimensional_NGRC}}.

While this study focused on NGRC, the insights presented here are broadly relevant to other data-driven modeling approaches relying on solving linear equations and least-squares problems. In particular, our findings resonate with similar concerns in Dynamic Mode Decomposition (DMD) \cite{Drmavc_2019,Pan_2020}, where matrix conditioning also plays a central role. {Furthermore, the conditioning of the feature matrix itself deserves deeper investigation.} Classical numerical analysis states that conditioning is highly sensitive to the distribution of points being evaluated (collocation points or nodes) and the choice of polynomial functions \cite{Kuian_2019}. For example, Vandermonde matrices evaluated at points nearly equally spaced on the unit circle or using orthogonal polynomial bases (like Chebyshev) are known to be well-conditioned. These properties, however, are unrelated to the underlying dynamics. This observation raises an exciting research direction of constructing a basis of orthogonal polynomials adapted to the distribution of points generated by the trajectory of the dynamical system. A first step toward this direction has been applied to coupled maps for network reconstruction \cite{Pereira_2025}. {Finally,} our results highlight the need for careful hyperparameter selection to ensure numerical stability of NGRC models. This numerical analysis is a first step toward designing better strategies to construct robust reservoir computers.

\section{Appendix}

\subsection{Metrics} 
\label{sec:metrics}
The reconstructed dynamics of the NGRC model is compared to the original dynamics using different metrics as detailed below.

\topic{Valid prediction time} We use the valid prediction time \cite{Vlachas_2020} to quantify the number of iterations such that the trajectories of the NGRC model and the original dynamics remain close to each other under a certain threshold. For a fixed threshold $\eta = 0.9$, we define the valid prediction time as 
\begin{align}
\begin{split}
    n_{VPT} &:= \min_n \Big\{ n : \frac{\|\x_n - \rs_n\|}{\sqrt{\frac{1}{N}\sum_{k = 0}^{N}\|\x_k\|^2}} > \eta\Big\} \\
    VPT &:= \frac{1}{T_\Lambda}n_{VPT} h,
\end{split}
\end{align}
where $T_{\Lambda} = \frac{1}{\Lambda}$ is the maximum Lyapunov exponent of the original dynamical system. For instance, for the Lorenz system, $\Lambda = 0.9056$.

\topic{Distance between successive maxima map} We introduce a distance between the induced successive maxima map of the original dynamics and the NGRC model. To this end, we utilize the successive local maxima from the time series of $z$-coordinate of the original and the NGRC model, $\{z_{n}^{\max}\}_{n \geq 0}$ and $\{ \hat{z}_{n}^{\max} \}_{n \geq 0}$, respectively. To compute a smooth approximation of the successive maxima maps:
\begin{align*}
S: z_{n}^{\max} \mapsto z_{n + 1}^{\max}, \quad 
\hat{S}: \hat{z}_{n}^{\max} \mapsto \hat{z}_{n + 1}^{\max},
\end{align*}
we use B-spline interpolation of degree $3$. Let $\mathcal{I} \subset \mathbb{R}$ be the overlapping domain of both maps, then we define the mean absolute difference over $1000$ evenly spaced points $\{u_j\}_{j=1}^{1000} \subset \mathcal{I}$ as:
\begin{align*}
D = \frac{1}{1000} \sum_{j=1}^{1000} |S(u_j) - \hat{S}(u_j)|.
\end{align*}

\topic{Power spectrum density} To capture the long time statistics of the trajectories we computed the power spectrum density. In the case the trajectories sample the attractor similarly over time, the power spectrum density should be close to other. The attractor similarity is computed using Kullback-Leibler (KL) divergence between the two empirical power spectrum densities of the trajectories. More precisely, from the trajectories $\{\x_n\}_{n}$ and $\{\rs_n\}_{n}$ we estimate the power spectrum density $P_{i}$ and $\hat{P}_{i}$ of each $i$th coordinate, respectively, using Welch's method \cite{Welch_1967} with Hann window, and compute the divergence error $E$ based on the KL divergence 
\begin{align*}
    D_{KL}(P_i||\hat{P}_i) &= \sum_{p = 0}^{\frac{L}{2} + 1} P_i(f_p) \log_{10}\Big(\frac{P_i(f_p)}{\hat{P}_{i}(f_p)}\Big), \\
    E &= \sum_{i = 1}^d D_{KL}(P_i||\hat{P}_i).
\end{align*}
where $L = \floor{\frac{5}{dt}}$ is the number of segments, the number of overlapping points $L/2$, computing the periodograms with mean.  

\section*{Supplementary Material}

{The Supplementary Material includes} a description of additional numerical experiments using the Double-Scroll electronic circuit.

\section*{Author Declarations}

\noindent
\textbf{Acknowledgments.} The authors thank Zheng Bian and Thomas Peron for enlightening discussions. ERS and EB thank the Collaborative Research in Computational Neuroscience (CRCNS) through R01-AA029926. The ONR, ARO, NIH, and DARPA also support EB.

\noindent
\textbf{Conflict of Interest.}
The authors have no conflicts to disclose.

\noindent
\textbf{Author Contributions.}
ERS: Conceptualization (equal); Formal analysis (lead); Investigation (lead); Methodology (lead); Software (lead); Visualization (lead); Writing – original draft (lead); Writing– review \& editing (equal). EB: Conceptualization (equal); Supervision (lead); Writing– review \& editing (equal); Funding acquisition (lead); Project administration (lead);

\section*{Data Availability Statement}
The code to produce the data supporting this study's findings is available in the \url{https://github.com/edmilson-roque-santos/numerical_stability_NGRC}. 

\bibliography{references} 


\clearpage

\newcommand\SupplementaryMaterials{%
  \xdef\presupfigures{\arabic{figure}}
  \xdef\presupsections{\arabic{section}}
  \renewcommand{\figurename}{Supplementary Figure}
  \renewcommand\thefigure{S\fpeval{\arabic{figure}-\presupfigures}}
  \renewcommand\thesection{S\fpeval{\arabic{section}-\presupsections}}
}

\SupplementaryMaterials

\clearpage
\onecolumngrid
\setcounter{page}{1}
\renewcommand{\thepage}{S\arabic{page}}

\begin{center}
{\large\bf Supplementary Material of}\\[5mm]
{\large{\textit{ On the emergence of numerical instabilities in Next Generation Reservoir Computing}}}\\[5pt]
Edmilson Roque dos Santos, and Erik Bollt.\\[5pt]
\end{center}
\section{Double-Scroll electronic circuit}

This supplementary material replicates the numerical experiments presented in the main text for the Lorenz-63 system, now applied to the Double-Scroll electronic circuit introduced in \cite{Chang_1998_DoubleScroll_intro}. The equations of motion in dimensionless form are given by
\begin{align}\label{eq:DS_equations_of_motion}
\begin{split}
    \dot{V}_1 &= \frac{V_1}{R_1} - \frac{\Delta V}{R_2} - 2 I_r \sinh{(\alpha \Delta V)}, \\
    \dot{V}_1 &= \frac{\Delta V}{R_2} + 2 I_r \sinh{(\alpha \Delta V)} - I, \\
    \dot{I} &= V_2 - R_4 I,
\end{split}
\end{align}
where $\Delta V := V_1 - V_2$. The parameters are the same used in \cite{Gauthier2021}: $R_1 =  1.2$, $R_2 = 3.44$, $R_4 = 0.193$, $\alpha = 11.6$, and $I_r = 2.25 \times 10^{-5}$, resulting in a Lyapunov time of $7.81$-time units. Because the equations in \cref{eq:DS_equations_of_motion} include non-polynomial terms, the NGRC model must approximate the original dynamics using polynomial basis functions. Numerical integration uses a fourth-order Runge-Kutta method with a fine time step $0.001$, and the time series is then subsampled at every time step $h$.

\begin{figure*}[t!]
\centering
\includegraphics[width=0.8\linewidth]{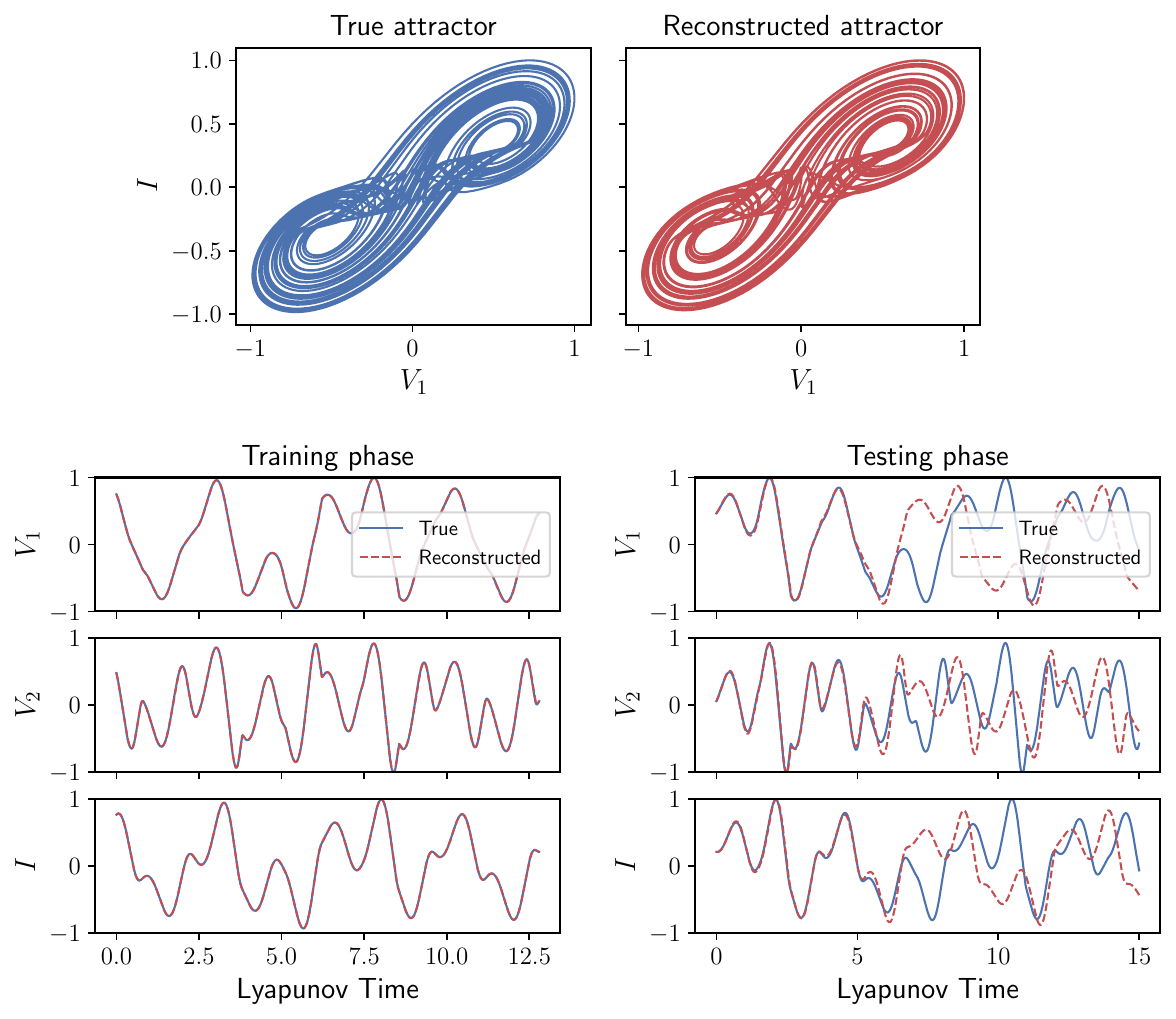}
\caption{\textbf{NGRC reproduces the Double-Scroll electronic circuit attractor.} The top panel shows the true (blue) and NGRC reconstructed (red) Double-Scroll attractor. The bottom panel displays NGRC performance during training and testing. The NGRC can forecast the original dynamics up to 5 Lyapunov times. The numerical integrator is the fourth-order Runge-Kutta with $h = 0.01$ and the solution is computed by the SVD algorithm. The parameters are delay dimension $k = 1$, time lag $\tau = 1$, the maximum degree $p = 3$, $\Ntrain = 10000$, $\Ntest = 80000$ and regularizer parameter $\beta = 0$.}
\label{fig:double_scroll}
 \end{figure*}

Consistent with the main text, we observe: NGRC accurately reproduces the Double-Scroll attractor without requiring regularization. \cref{fig:double_scroll} shows the attractor reconstruction and forecasting capability of the NGRC model for the Double-Scroll system. For this example, the condition number $\kappa(\Psi) = 1198$ and the closeness of fit for the three coordinates is reported in \cref{table:closeness_of_fit_supp}. All algorithms achieve nearly identical closeness of fit up to machine precision. More importantly, the values are much larger than the order $10^{-14}$ as observed for the Lorenz system, primarily due to the $\sinh{}$ term,  requiring that polynomial approximation (in this case of degree $3$), inevitably decreasing the fitting accuracy. Since the condition number remains well below the inverse of machine precision, the feature matrix is not in the ill-conditioned regime. As expected, the difference $\Delta$ between algorithms' solutions is small, on the order of $10^{-12}$, consistent with the discussion in the main text.

\begin{table}[ht]
\centering
\begin{tabular}{|c|c|c|c|}
\hline 
Algorithm & $\theta_{V_1}$ & $\theta_{V_2}$ & $\theta_{I}$ \\
\hline \hline
Cholesky  & $7.765499763260145 \times 10^{-2}$ & $9.133421534662849 \times 10^{-2}$ & $2.983845874175234 \times 10^{-4}$ \\  \hline
SVD & $7.765499763260143 \times 10^{-2}$ & $9.13342153466285 \times 10^{-2}$ & $2.98384587417524 \times 10^{-4}$ \\ \hline
LU & $7.765499763260143 \times 10^{-2}$ & $9.13342153466285 \times 10^{-2}$ & $2.9838458741752435 \times 10^{-4}$ \\\hline
\end{tabular}
\caption{Closeness of fit for each coordinate of the Double-Scroll system. The parameters are the same as those used for \cref{fig:double_scroll}.}
\label{table:closeness_of_fit_supp}
\end{table}

\subsection{Sparsely-sampled training data}
\begin{figure*}[t!]
\centering
\includegraphics[width=0.9\linewidth]{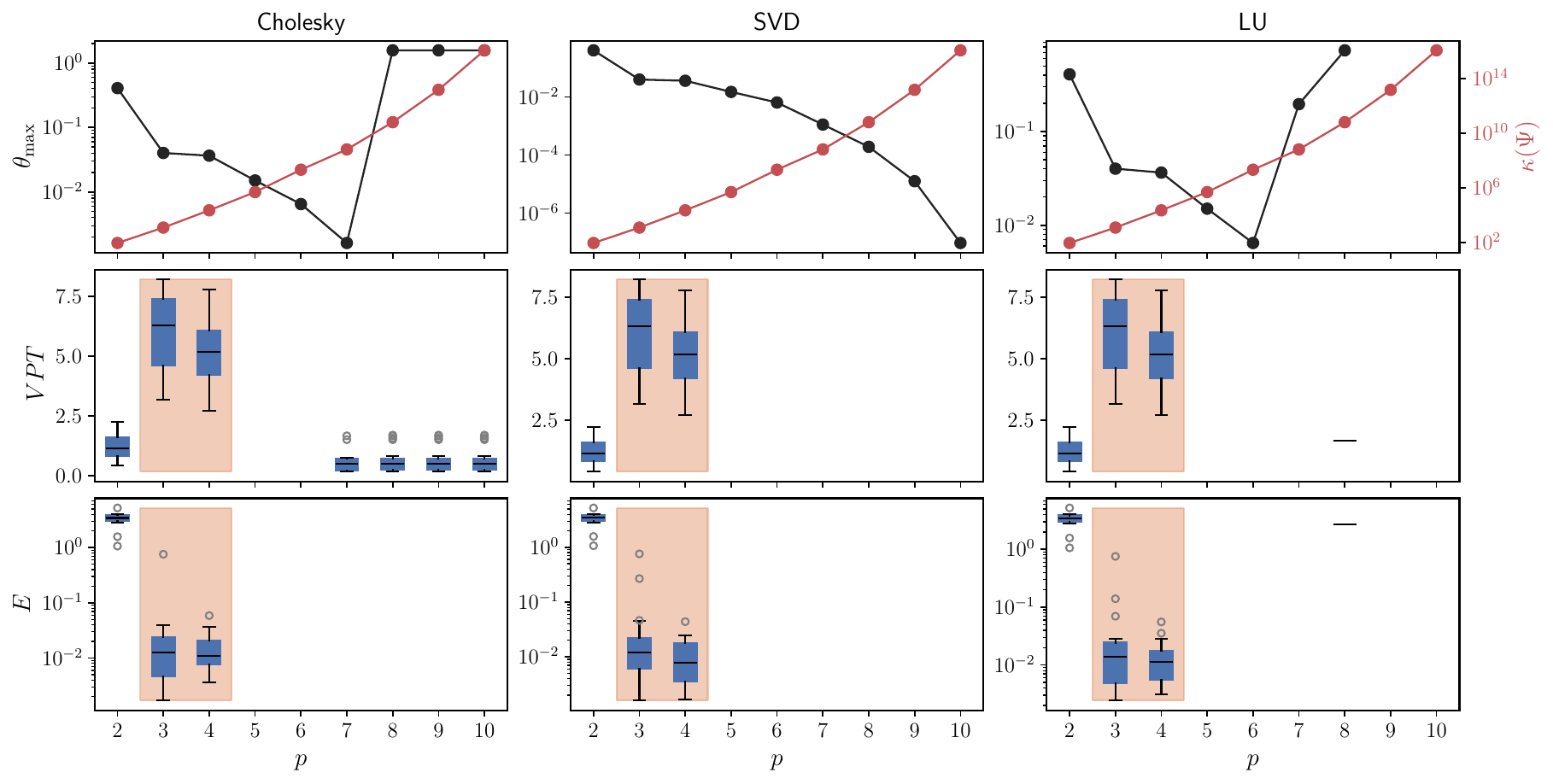}
\caption{\textbf{NGRC reproduces the original Double-Scroll dynamics using third-degree polynomial expansions.} The top panel shows the maximum closeness of fit over the three components (black) and the condition number (red) with respect to the maximum degree $p$. Subsequent panels show box plots over $25$ initial conditions. The orange shaded area shows the maximum degree where all three algorithms perform equally within statistical significance. The parameters are $h = 0.25$, delay dimension $k = 1$, time lag $\tau = 1$, $\Ntrain = 400$, $\Ntest = 6000$ and regularizer parameter $\beta = 0$.}
\label{fig:subsampling_solvers_DS}
 \end{figure*}

Consider a sparsely-sampled time series scenario for $h = 0.25$, similar to the main text, to assess the NGRC performance as the maximum degree $p$ increases, see \cref{fig:subsampling_solvers_DS}. The findings closely parallel the Lorenz system results: the condition number increases exponentially for increasing $p$, while the closeness of fit $\theta_{\max}$ behaves smoothly only for the SVD algorithm. For Cholesky and LU algorithms, the closeness of fit breaks down at $p = 6$, and becomes non-monotonic.

The NGRC performance is quantified by the valid prediction time and the divergence error $E$. NGRC reproduces accurately the attractor without the need for regularization for maximum degrees within the orange band. Unlike the Lorenz system, all three algorithms perform comparably in this regime, with results statistically similar, but degrade for $p > 4$. This difference is mostly due to the Double-Scroll system's non-polynomial vector field. While the Lorenz dynamics can be well-approximated by larger polynomial expansions due to the polynomial structure, the Double-Scroll dynamics is inherently harder to approximate. It is therefore surprising that a third-degree polynomial expansion suffices for useful forecasting. 

\subsection{Partial measurement}

\begin{figure*}[t!]
\centering
\includegraphics[width=0.55\linewidth]{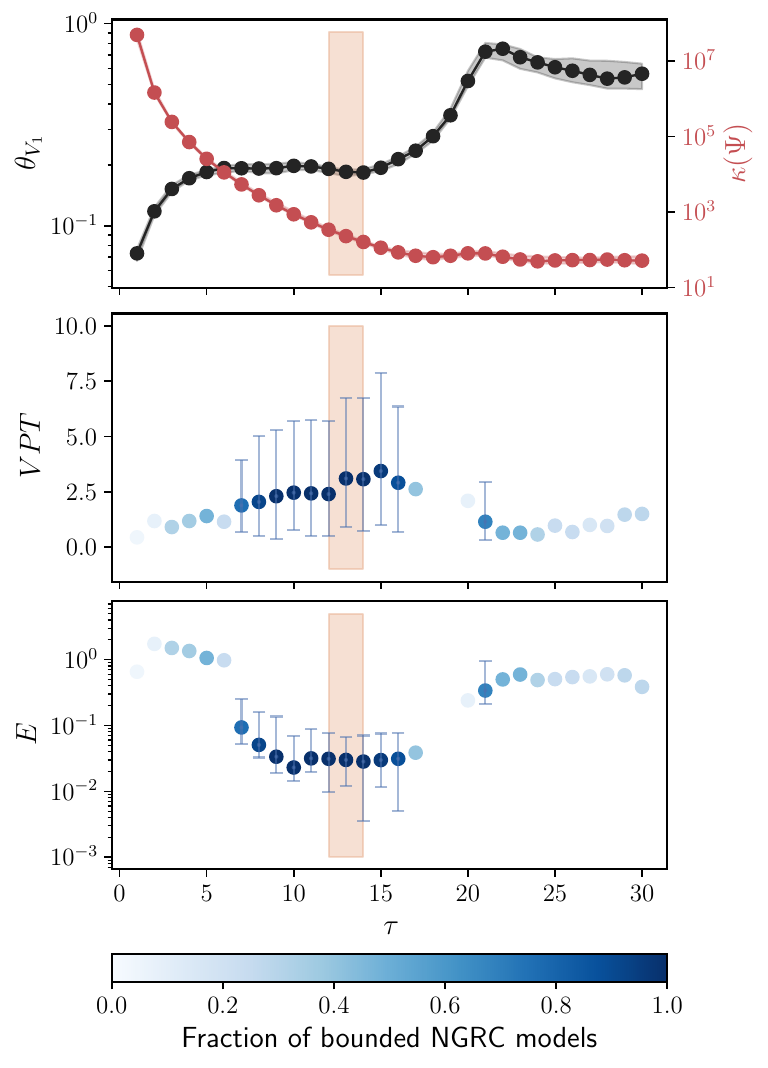}
\caption{\textbf{Methods used for embedding time series inform NGRC modelling of partial measurement of Double Scroll electronic circuit.} The top panel shows the closeness of fit $\theta_{V_1}$ (black) and condition number of $\Psi$ (red) with respect to the increasing time lag $\tau$. The subsequent panels show the NGRC performance in both metrics, VPT and divergence error $E$. The orange band represents the $\tau = 13$, which is informed by the first minimum of mutual information. The blue colormap represents the fraction of bounded NGRC models over the $25$ initial conditions during testing. The dots correspond to the median, and error bars are $25\%$ (lower bound) and $75\%$ (upper bound) quantiles, which are shown only for NGRC models with a fraction larger than $0.5$. The numerical integrator is the fourth-order Runge-Kutta method. The parameters are $h = 0.25$, delay dimension $k = 3$, maximum degree is $p = 3$, $\Ntrain = 400$, $\Ntest = 6000$ and regularizer parameter $\beta = 0$.}
\label{fig:partial_measurement_DS}
 \end{figure*}

We consider the partial measurement consisting solely of the $V_1-$coordinate time series. For $h = 0.25$, we choose delay dimension $k = 3$ (identified by false nearest neighbors method), based on  \cref{fig:subsampling_solvers_DS}, we set maximum degree $p = 3$. \cref{fig:partial_measurement_DS} shows that by increasing the time lag reduces the condition number monotonically, consistent with the results reported in the main text. The closeness of fit $\theta_{V_1}$ exhibits similar behavior as the Lorenz system case, with the difference that, for $\tau > 20$, the accuracy decays slightly instead of reaching a plateau. 

\begin{figure*}[t!]
\centering
\includegraphics[width=1.0\linewidth]{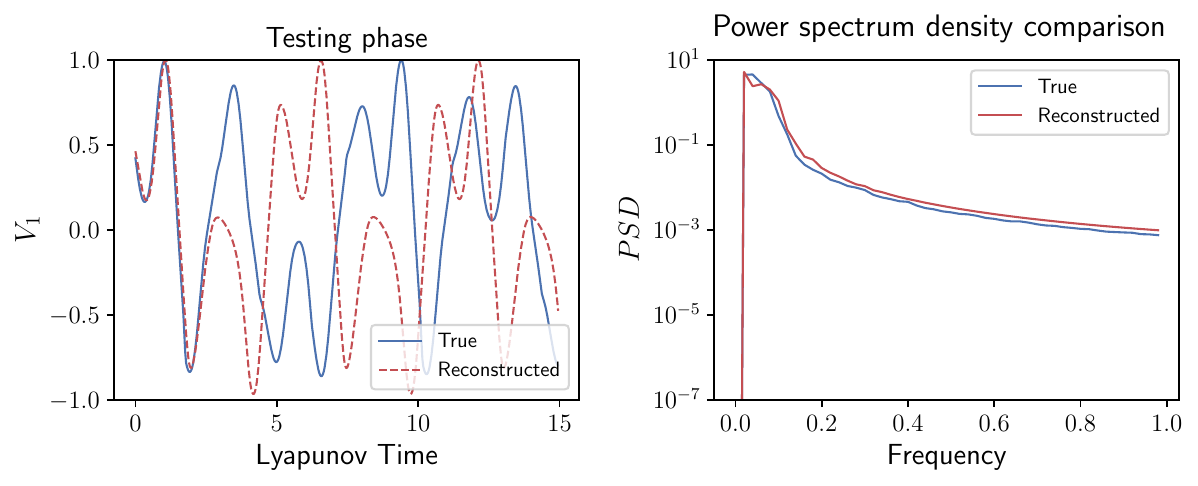}
\caption{\textbf{NGRC forecasts the partial measurement of the Double-Scroll.} The left panel shows the NGRC trajectory (red) compared to the original (blue). The right panel shows the comparison between the power spectrum densities. The parameters are $h = 0.25$, delay dimension $k = 3$, time lag $\tau = 13$, the maximum degree $p = 3$, $\Ntrain = 400$, $\Ntest = 6000$ and regularizer parameter $\beta = 0$.}
\label{fig:x_coord_DS}
 \end{figure*}

The NGRC performance is quantified by the valid prediction time, divergence $E$ between the power spectrum densities, and the fraction of bounded NGRC models. Interestingly, the results reveal a wide range of time lags yielding bounded models with statistically similar performance. Although the underlying reason here is unclear, this wide range contrasts with the Lorenz system (Runge–Kutta case), for which there was a narrow range of time lags. Importantly, this range includes the time lag suggested by the first minimum of mutual information method, $\tau = 13$, reinforcing that methods used for embedding time series can effectively guide hyperparameter choices. Using these informed hyperparameters, \cref{fig:x_coord_DS} compares the original and reconstructed time series, along with their power spectra. The NGRC model successfully forecasts approximately two Lyapunov times of the original trajectory.

\end{document}